\documentclass{article} % For LaTeX2e
\usepackage{arxiv}
\usepackage{booktabs}
\usepackage{amsmath,amssymb,bm,comment,environ,grffile,latexsym,verbatim,xspace}

\makeatletter
\@ifpackageloaded{float}{}{
  \usepackage{floatrow}
  \floatsetup[table]{capposition=top}
}
\makeatother

\makeatletter
\@ifpackageloaded{cite}{}{
  \@ifpackageloaded{natbib}{}{\usepackage[sort&compress,numbers]{natbib}}
}
\makeatother
\makeatletter
\@ifpackageloaded{algorithmic}{}{\usepackage[noend]{algpseudocode}}
\@ifpackageloaded{algorithm}{}{\usepackage[ruled]{algorithm}}
\makeatother

\usepackage{graphicx} % more modern
\usepackage{grffile} % support suffix

\makeatletter
\@ifclassloaded{beamer}{
  \usepackage[compatibility=false]{caption}
  }{
  \usepackage{caption}
}
\makeatother

\usepackage[labelformat=simple]{subcaption}
\usepackage{multirow}

\usepackage{tikz}
\usepackage{pgfplots}
\usetikzlibrary{shapes,shapes.arrows,shapes.gates.logic.US,trees,positioning,arrows,automata,decorations.pathmorphing,chains}
\usetikzlibrary{shadows,calc,backgrounds,patterns,fit}
\usetikzlibrary{matrix,arrows}
\usepackage{url}

\makeatletter
\@ifpackageloaded{hyperref}{}{
  \usepackage[colorlinks,linkcolor={red!80!black},citecolor={blue!80!black},urlcolor={blue!80!black}]{hyperref}
}

\@ifclassloaded{beamer}{
  \usepackage{amsfonts}
  }{
  \usepackage{times}
  \usepackage{mathrsfs,dsfont} % dsfont to support
  \usepackage{mathbbol}
  \DeclareSymbolFontAlphabet{\mathbbl}{bbold}
  \usepackage{amsfonts}
  \DeclareSymbolFontAlphabet{\mathbb}{AMSb}
  \usepackage{enumitem}
  \setlist{noitemsep}
  \setlist[1]{noitemsep}
  \setlist[1]{nosep}

  \newlist{compactitem}{itemize}{3}
  \setlist[compactitem]{topsep=0pt,partopsep=0pt,itemsep=0pt,parsep=0pt,leftmargin=*}
  \setlist[compactitem,1]{label=\textbullet}
  \setlist[compactitem,2]{label=---}
  \setlist[compactitem,3]{label=*}

  \newlist{compactdesc}{description}{3}
  \setlist[compactdesc]{topsep=0pt,partopsep=0pt,itemsep=0pt,parsep=0pt}

  \newlist{compactenum}{enumerate}{3}
  \setlist[compactenum]{topsep=0pt,partopsep=0pt,itemsep=0pt,parsep=0pt}
  \setlist[compactenum,1]{label=\arabic*}
  \setlist[compactenum,2]{label=\alph*}
  \setlist[compactenum,3]{label=\roman*}
}
\makeatother

\usepackage{etex}  % For larger registers for latex
\usepackage{color}
\usepackage{mathtools}
\usepackage{ifthen}
\usepackage{etoolbox}

\usepackage{xr}
\makeatletter
\newcommand*{\addFileDependency}[1]{
        \typeout{(#1)}
        \@addtofilelist{#1}
        \IfFileExists{#1}{}{\typeout{No file #1.}}
}
\makeatother

\input{defs.tex}

\newcommand{\best}[1]{{\bf #1}}

\def\liblinear{LIBLINEAR\xspace}
\def\pecos{PECOS\xspace}
\def\xmc{XMR\xspace}
\def\xmclong{eXtreme Multilabel Ranking\xspace}

\def\xtransformer{X-TRANSFORMER\xspace}
\def\xrlinear{XR-LINEAR\xspace}
\def\xrtransformer{XR-TRANSFORMER\xspace}

\newcommand{\attentionxml}{AttentionXML\xspace}
\newcommand{\annexml}{AnnexML\xspace}
\newcommand{\bonsai}{Bonsai\xspace}
\newcommand{\discmec}{DiSMEC\xspace}

\newcommand{\fasttext}{fastText\xspace}

\newcommand{\parabel}{Parabel\xspace}
\newcommand{\pfastrexml}{PfastreXML\xspace}

\newcommand{\proxml}{ProXML\xspace}

\newcommand{\slice}{SLICE\xspace}
\newcommand{\xmlcnn}{XML-CNN\xspace}
\newcommand{\xtext}{eXtremeText\xspace}
\newcommand{\napkinxc}{NAPKINXC\xspace}
\newcommand{\lightxml}{LightXML\xspace}

\newcommand{\deepxml}{DeepXML\xspace}
\newcommand{\decaf}{DECAF\xspace}
\newcommand{\galaxc}{GalaXC\xspace}
\newcommand{\eclare}{ECLARE\xspace}

\newcommand{\pifa}{{\sf PIFA}\xspace}
\newcommand{\pii}{{\sf PII}\xspace}
\newcommand{\spectral}{{\sf Spectral}\xspace}
\newcommand{\lf}{{\sf LF}\xspace}
\newcommand{\elmo}{{\rm ELMo}\xspace}
\newcommand{\tfidf}{tfidf\xspace}

\newcommand{\eurlex}{{\sf Eurlex-4K}\xspace}
\newcommand{\wikis}{{\sf Wiki10-31K}\xspace}
\newcommand{\amzcat}{{\sf AmazonCat-13K}\xspace}
\newcommand{\amzsmall}{{\sf Amazon-670K}\xspace}
\newcommand{\amzlarge}{{\sf Amazon-3M}\xspace}
\newcommand{\wikil}{{\sf Wiki-500K}\xspace}

\setlist[compactitem]{leftmargin=*}

\title{PECOS: \textbf{P}rediction for \textbf{E}normous and \textbf{C}orrelated \textbf{O}utput \textbf{S}paces}
\date{}
\author{
  {Hsiang-Fu Yu} \\
	Amazon\\
	\texttt{rofu.yu@gmail.com} \\
	\And
  {Kai Zhong} \\
	Amazon\\
  \texttt{kaizhong89@gmail.com} \\
	\And
  {Jiong Zhang} \\
	Amazon\\
  \texttt{zhangjiong724@gmail.com} \\
	\And
  {Wei-Cheng Chang} \\
	Amazon\\
  \texttt{weicheng.cmu@gmail.com} \\
	\And
  {Inderjit S. Dhillon} \\
	UT Austin \& Amazon\\
	\texttt{inderjit@cs.utexas.edu}
}

\begin{document}
\maketitle

\begin{abstract}%
Many large-scale applications amount to finding relevant results from an enormous output space of potential candidates.
For example, finding the best matching product from a large catalog or suggesting related search phrases on a search engine.
The size of the output space for these problems can range from millions to billions, and can even be infinite in some applications.
Moreover, training data is often limited for the ``long-tail'' items in the output space.
Fortunately, items in the output space are often correlated thereby presenting an opportunity to alleviate the data sparsity issue.
In this paper, we propose the Prediction for Enormous and Correlated Output Spaces~(\pecos) framework,
a versatile and modular machine learning framework for solving prediction problems for very large output spaces,
and apply it to the \xmclong~(\xmc) problem:
given an input instance, find and rank the most relevant items from an enormous but fixed and finite output space.
We propose a three phase framework for \pecos:
(i)~in the first phase, \pecos organizes the output space using a semantic indexing scheme,
(ii)~in the second phase, \pecos uses the indexing to narrow down the output space by orders of magnitude using a machine learned matching scheme, and
(iii)~in the third phase, \pecos ranks the matched items using a final ranking scheme.
The versatility and modularity of \pecos allows for easy plug-and-play of various choices for the indexing, matching, and ranking phases.
The indexing and matching phases alleviate the data sparsity issue by
leveraging correlations across different items in the output space.
%thereby strengthening statistical signals and allowing \pecos to make better quality predictions.
For the critical matching phase, we develop a recursive machine learned matching strategy with both linear and neural matchers.
When applied to \xmclong where the input instances are in textual form,
we find that the recursive Transformer matcher gives state-of-the-art accuracy results,
at the cost of two orders of magnitude increased training time compared to the recursive linear matcher.
For example, on a dataset where the output space is of size 2.8 million,
the recursive Transformer matcher results in a 6\% increase in precision@1 (from 48.6\% to 54.2\%)
over the recursive linear matcher but takes 100x more time to train.
Thus it is up to the practitioner to evaluate the trade-offs
and decide whether the increased training time and infrastructure cost is warranted for their
application; indeed, the flexibility of the \pecos framework seamlessly allows
different strategies to be used. We also develop very fast inference
procedures which allow us to perform \xmc predictions in real time; for
example, inference takes less than 1 millisecond per input on the dataset with
2.8 million labels.
The \pecos software is available at \url{https://libpecos.org}.
\end{abstract}

\section{Introduction}

\label{sec:xmc-formulation}

Many challenging problems in modern applications amount to finding relevant
results from an enormous output space of potential candidates, for example,
finding the best matching product from a large catalog or suggesting related
search phrases on a search engine. The size of the output space for these
problems can range from millions to billions, and can also be infinite in some
applications. For example, when suggesting related searches, the output space
is the set of valid search phrases which is clearly infinite; many valid
search phrases have already been seen by the search engine, but on emerging
topics a new search phrase may need to be synthesized.

\begin{figure}[!t]
    \centering
    \includegraphics[width=0.8\linewidth]{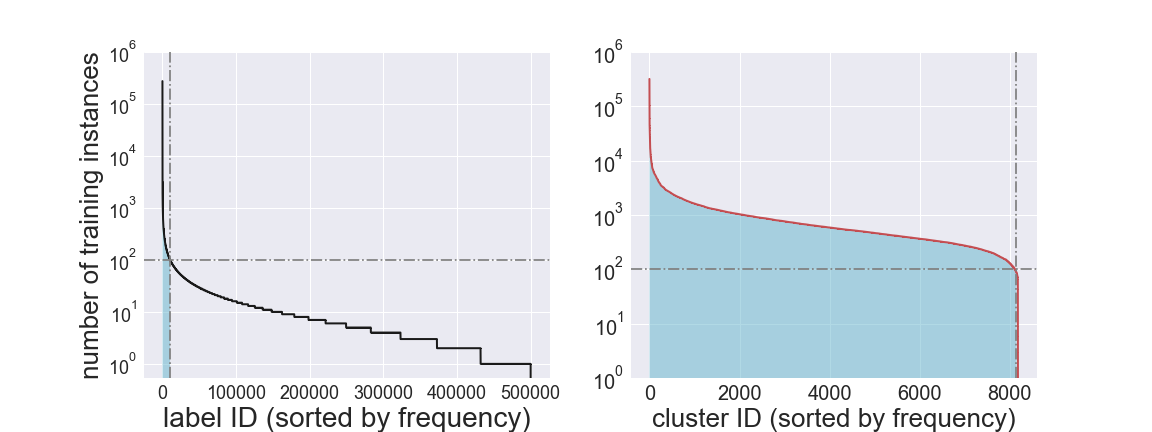}
    \caption{On the left, \wikil shows a long-tail distribution of training data over the labels.
      Only $2.1\%$ of the labels have more than 100 training instances, as
      indicated by the cyan blue regime.  On the right is the
      distribution of labels after our semantic label indexing is performed to
      form 8,192 label clusters; $99.4\%$ of the clusters have more than 100 training
      instances, which mitigates the data sparsity issue.
    }
    \label{fig:label-distr-wiki}
\end{figure}

% label sparsity
Moreover, observational or training data is often limited for many of the
so-called ``long-tail'' of items in the output space. Given the inherent
paucity of training data for most of the items in the output space, developing
machine learned models that perform well for spaces of this size is
challenging. We illustrate these challenges on a multi-label problem, where
the goal is to assign or predict labels for a new input instance.  Consider
the \wikil dataset~\citep{xmc_repo}, where the problem is to assign text
labels to a Wikipedia page from a known label set. The left panel of
Figure~\ref{fig:label-distr-wiki} shows that only $2\%$ of the labels have
more than 100 ``positive'' training instances, while the remaining $98\%$ are
``long-tail'' labels with many fewer training instances. Due to this severe
data sparsity issue, it is challenging to design an effective multilabel
strategy that assigns tail labels to input instances.

Fortunately, items in the output space are often correlated thereby presenting
the opportunity to alleviate the data sparsity issue. In this paper, we
exploit these correlations in the output space and propose the Prediction for
Enormous and Correlated Output Spaces~(\pecos) framework, a versatile
and modular machine learning framework for solving prediction problems for
very large output spaces, and apply it to the \xmclong~(\xmc)
problem: given an input instance, find and rank the most
relevant items from an enormous but fixed and finite output space. We propose
a three phase framework for \pecos:
(i) in the first phase, \pecos organizes the output
space using a semantic indexing scheme, (ii) in the second phase, \pecos uses
the indexing to narrow down the output space by orders of magnitude using a
machine learned matching scheme, and (iii) in the third phase, \pecos ranks the
matched items using a final ranking scheme. The indexing and matching phases
alleviate the data sparsity issue by leveraging correlations across different
items in the output space thereby strengthening statistical signals. As an
example, consider again the \wikil dataset where \pecos performs semantic
indexing by clustering the labels; on the right panel of
Figure~\ref{fig:label-distr-wiki} we show the distribution of training data
over the label clusters. Now, over 99\% of label clusters have more than 100
training instances; this allows transfer of training signals to tail items and
alleviates the data sparsity issue thus allowing \pecos
to make better quality predictions.

For the critical matching phase,
we investigate a recursive machine learned linear matching strategy
as well as a recursive deep learned neural matcher.
When applied to \xmclong where the input instances are in textual form,
we find that the recursive neural matcher based on Transformer encoders gives state-of-the-art accuracy results;
however the recursive Transformer matcher requires about two orders of magnitude increased training time than the recursive linear matcher.
%On the other hand, the recursive linear matching strategy is much more efficient in both training as well as inference, but yielding slightly lower quality than the neural matcher.
%For example, on the~\amzcat data set, the neural matcher results in a 192 basis points (bps) increase in precision@1 but at the additional cost of $\$3,100$ dollars in training, resulting in an incremental cost of  $3,100/192\$=16.15$ dollars per 1 bps improvement in the desired precision metric. If training is done daily, as is common in many industrial applications, this would amount to an increased cost of \$ 1.1 million per year (the increase in inference costs would be extra).
For example, on a dataset where the output space is of size 2.8 million,
the recursive Transformer matcher results in a 6\% increase in precision@1 (from 48.6\% to 54.2\%)
over the recursive linear matcher but takes 100x more time to train.
%If training is done daily, as is common in many industrial applications, this would amount to an increased cost of \$3 million per year~(the increase in inference costs would be extra).
Thus it is up to the practitioner to evaluate the trade-offs
and decide whether the increased training time and infrastructure cost is warranted for their
application; indeed, the flexibility of the \pecos framework seamlessly allows
different strategies to be used.

Our contributions in this paper are summarized as follows:
\begin{itemize}
  \item We propose the Prediction for Enormous and Correlated Output
    Spaces~(\pecos) framework, a versatile and modular machine learning framework
    for solving prediction problems for very large output spaces. The
    versatility of \pecos comes from our proposed three-phase approach: (i)
    semantic label indexing, (ii) machine learned matching, and finally (iii)
    ranking.
  \item The flexibility of \pecos allows practitioners to evaluate the
    trade-offs between performance and infrastructure cost to identify the
    most appropriate \pecos variant for their application.
  \item To exhibit the flexibility of \pecos, we propose three concrete
    realizations:
    (i) \xrlinear is a recursive linear machine learned realization of our \pecos framework;
    (ii) \xtransformer is a neural realization of the \pecos framework \textit{without recursive} Transformer encoders;
    (iii)\xrtransformer is a neural realization of the \pecos framework \textit{with recursive} Transformer encoders;
    %yields better prediction accuracy metrics than \xrlinear but needs much larger training time.
  \item We present detailed experimental results showing that \pecos yields
    {\em state-of-the-art} results for \xmc in terms of precision, recall, and
    computational time (training and inference).
\end{itemize}
Parts of this paper related to \xtransformer and \xrtransformer
have appeared in~\citet{chang2020xmctransformer} and ~\citet{zhang2021fast}, respectively,
while the part related to fast inference has appeared in~\citet{etter2021accelerating}.

This paper is organized as follows: we setup the problem formulation in Section~\ref{sec:intro-scene}.
In Section~\ref{sec:pecos-framework}, we propose a three-phase framework for \pecos and describe each phase in detail.
Next, we present \xrlinear, a recursive linear machine learned realization in Section~\ref{sec:xrlinear},
and \xtransformer and \xrtransformer, the neural realization for inputs in textual form in Section~\ref{sec:deep-matcher}.
We then discuss the connections of \pecos to related work in Section~\ref{sec:related-work}.
We present detailed experimental results in Section~\ref{sec:exp} and conclude our paper in Section~\ref{sec:future}.
The \pecos software is available at \url{https://libpecos.org}.

\subsection{Setting the Scene}
\label{sec:intro-scene}
% define \xmc problem
In this paper, we focus on the \xmclong~(\xmc)
problem: given an input instance, return the most relevant labels from an
enormous label collection, where the number of labels could be in the millions
or more.
% General form of \xmc problem
One can view the \xmc problem as learning a score function
$f: \cX \times \cY \rightarrow \RR$, that maps an (instance, label) pair
$(\bx, \ell)$ to a score $f(\bx, \ell)$, where $\ell\in\cY$.  The function $f$ should
be optimized such that highly relevant $(\bx, y)$ pairs have high scores,
whereas irrelevant pairs have low scores.
% Many applications
Many real-world applications are in this form.  For example, in E-commerce
dynamic search advertising, $\bx$ represents an item and $\ell$ represents a
bid query on the market~\citep{prabhu2014fastxml,prabhu2018parabel}.  In
open-domain question answering, $\bx$ represents a question and $\ell$
represents an evidence passage containing the
answer~\citep{lee2019latent,chang2020pretraining}.  In the PASCAL Large-Scale
Hierarchical Text Classification (LSHTC) challenge, $\bx$ represents an
article and $\yb$ represents a category in the  hierarchical Wikipedia
taxonomy~\citep{partalas2015lshtc}.

Formally speaking, in an \xmc problem, we are given a training dataset
$\cbr{(\bx_i, \by_i):i=1,\ldots,n}$, where $\bx_i \in \RR^d$ is a $d$-dimensional
feature vector for the $i$-th
instance, and $\by_i \in \cbr{0,1}^L$ denotes the relevant labels for this
instance from an output space $\cY \equiv \cbr{1,\ldots,\ell,\ldots,L}$ with $L$
labels. In a typical \xmc problem, the number of instances $n$, the number of
instance features $d$, and the number of labels $L$ can all be in the
millions, or larger. We
also use $X=[\bx_1,\ldots,\bx_i,\ldots,\bx_n]^\top \in \RR^{n \times d}$ to
denote the input feature matrix, and use
$Y=[\by_1,\ldots,\by_i,\ldots,\by_n]^\top \in \cbr{0,1}^{n\times L}$ to denote
the input-to-label matrix.

The scoring function $f(\bx, \ell): \RR^d\times \cY \rightarrow \RR$ is to be
learned from the given training data such that $f(\bx, \ell)$ maps an input
(or instance) $\bx$ and a label $\ell$ to a relevance score, which can be used
to identify labels most relevant to $\bx$ from the output space $\cY$. Since
the top scores are the salient ones, we further use $f_b(\bx)\subset \cY$ to
denote the top-$b$ predicted labels for a given instance $\bx$, i.e.,
\[
  f_b(\bx) = \mathop{\arg\max}_{\cS\subset\cY:\abs{\cS}=b} \sum_{\ell \in \cS} f(\bx, \ell).
\]

To evaluate the feasibility of $f(\cdot)$ for a given \xmc problem, we need to
consider the following questions:
\begin{itemize}
  \item Quality of $f_b(\cdot)$: how well does $f_b(\cdot)$ perform on an unseen input instance $\bx$?
  \item Training Efficiency: how efficient is the training
    algorithm in learning the parameters of $f(\cdot)$?
  \item Inference Speed: how fast is the computation of $f_b(\cdot)$ to serve real-time requests?
  \item Infrastructure Cost: how much do the training and inference procedures cost in terms
    of computational resources?
\end{itemize}

As an example, using the vanilla linear one-versus-rest (OVR) approach~\citep{bishop2006pattern}, the
scoring function can be defined on each label as follows:
\begin{align}
  f(\bx, \ell) = \bw_{\ell}^\top\bx,\ \forall \ell \in \cY.
  \label{eq:ova-linear}
\end{align}
The parameter $\bw_{\ell} \in \RR^d$ for the $\ell$-th label may be  obtained by solving a
regularized binary classification problem:
\begin{align}
  \bw_{\ell} = \arg\max_{\bw\in \RR^d} \sum_{i=1}^n \cL(Y_{i\ell}, \bw^\top \bx_i) + \frac{\lambda}{2}
  \bw^\top\bw,
  \label{eq:binary-erm}
\end{align}
where $\cL(\cdot, \cdot)$ is a loss function and $\lambda>0$ is a
regularization hyperparameter.  The overall parameter space for this linear OVR
approach is $\cO(dL)$, the training time is
$L \times T_{\text{binary}}$, and the inference time
is $\Omega(dL)$, where $T_{\text{binary}}$ is the time required to train a
binary classifier. Usually, $T_{\text{binary}}$ is at least
linear in  the number of nonzeros in the training data, which we denote by $\text{nnz}(X)$.

Let us sketch a ballpark estimate of how long it would take to train an OVR
model on the \wikil dataset with $n=$ 1.5 million text training
documents and $L=$ 0.5 million output labels, with average number of tokens
in each document being around $1,000$.
Further suppose that we use logistic regression as our binary classifier and the term
frequency-inverse document frequency (\tfidf) vectorizer with a vocabulary of
size 2.5 million to form a sparse training matrix $X\in\RR^{1.5 \cdot 10^6\times 2.5\cdot 10^6}$.
With such a training matrix $X$
each logistic regression can be trained in about 50
seconds~\citep{fan2008liblinear,hsieh2008dual}.\footnote{Our 50 second estimate is obtained by
extrapolating the running time in terms of nonzero entries from~\citet[Table 2]{hsieh2008dual}.}
Note that the training time remains similar even if
we use dense embeddings to form a dense feature matrix $X$ as  $\text{nnz}(X)$
might be even larger than the sparse \tfidf feature matrix.
In this setting, the overall training time will be around
$1.5 \cdot 10^6 \times 50 = 7.5 \cdot 10^7 \text{ seconds } =
\frac{7.5 \cdot 10^7}{\rbr{60 \times 60 \times 24 \times 30}} \approx 29 \; \text{months}$
on 1 CPU and 1.8 months even with perfect 16-way parallelization! Further, the
full model would require a prohibitive amount of
$2.5\cdot 10^6\times 0.5 \cdot 10^6\times 4 \; \text{Bytes} \approx $ $5$ TB disk
space (assuming a single precision floating point format).  Clearly
such a simple approach is not feasible for an \xmc problem of this
scale even with linear models.
Note that the simple deep learning extension to multi-label classification
(e.g., binary cross entry or softmax losses), which
jointly learns parameters for all labels, only imposes more computational
requirements. Another important aspect is that
the inference for the OVR approach is extremely slow
when $L$ is large as the inference time has $\cO(L)$ complexity. Clearly, both
training and inference times and memory are prohibitive for a simple OVR
solution. In contrast, with \pecos, we are able to train a dataset of this
size ($n=1.5\cdot 10^6$, $d=2.5\cdot 10^6$, $L=0.5\cdot10^6$) in 5 hours on a 16 CPU
machine with the model requiring about 5 GB space. This \pecos model also allows inference in $\cO(\log L)$ time.

\section{\pecos \xmc Framework}
\label{sec:pecos-framework}
In order to build a framework to solve general \xmc problems, we borrow from the
design of modern information retrieval (IR) systems
where the goal is to find the top few relevant documents for a given query from an
extremely large number of documents. IR can be regarded as a special
\xmc problem with queries as inputs and documents as output labels. Furthermore,
when both queries and documents are in the same text domain, an efficient and
scalable IR system such as Apache Lucene\footnote{https://lucene.apache.org/},
typically consists of the following stages~\citep{how_search_work}; 1)
{lexical indexing}: building an efficient data structure in an offline
manner to lexically index the documents by its tokens; 2) {lexical matching}: finding the
documents that contain this query; and 3) {ranking}: scoring the matched
documents, often using machine learning. This three stage design is crucial to any scalable IR system
that deals with a large number of documents.

Albeit scalable, an IR system cannot be easily
generalized to handle general \xmc problems due to the following reasons.
First, output labels for a general \xmc problem might not have text information
so the lexical indexing approach is not applicable.  Second, input instances
and output labels for a general \xmc problem might not be in the same domain, for
example, the inputs could be images and the labels could be image annotations.
Third, input instances for a general \xmc problem are usually in a
feature vector form instead of in text form, so the indexing/matching
techniques in IR are not applicable.

Motivated by the design for IR systems, in \pecos, we propose a three-stage
framework to solve general \xmc problems in a scalable and modular manner:
\begin{itemize}
  \item { Semantic Label Indexing}: we organize the original label space $\cY$,
    $\abs{\cY} = L$,
    so that semantically similar labels are arranged together.
    One way is to partition the labels into $K$ clusters,
    $\cbr{\cY_k: k=1,\ldots,K}$, where $K \ll L$ and each cluster $\cY_k$ is a subset of
    labels which are ``semantically similar'' to each other. Similar to
    lexical indexing in IR, semantic label indexing is constructed in an
    offline manner before the training is done. Alternatives for semantic
    indexing are an approximate kNN index structure.
  \item { Machine-learned Matching}: we learn a scoring function $g(\bx, k)$
    that maps an input $\bx$ to relevant indices/clusters denoted
    by an indicator vector $\bmhat \in \cbr{0,1}^{K}$, where the $k$-th
    element $\mhat_k=1$ denotes that the $k$-th index/cluster is deemed to be
    relevant for the input $\bx$.
  \item { Ranking}: we train a ranker $h(\bx, \ell)$ to give final
    scores of candidate labels shortlisted by the matcher (given by
    $\bigcup\limits_{k:\mhat_k=1} \cY_k$). The efficiency of the
    method stems from only ranking $\cO(K)$ ($\ll L$) labels.
\end{itemize}

With this three-stage framework, the inference complexity can be greatly
reduced. For a given input $\bx$, let $b=\nnz\rbr{\bmhat}$ be the number of
matched indices/clusters given by $g(\bx, k)$ and
let $\text{avg}\rbr{\abs{\cY_k}} = L/K$ be the average number of labels in each
index/cluster. The inference complexity is reduced to
\[
  \underbrace{
     K \times \cO(\text{time to evaluate } g(\bx, k))
   }_{\text{matcher time}} +
   \underbrace{
     b \times \frac{L}{K}
     \times \cO(\text{time to evaluate } h(\bx, \ell))
   }_{\text{ranker time}}.
\]
With an appropriate choice of $K$
and $b$, the overall inference time complexity can be  drastically reduced. For
example, assuming linear models that have time complexity $\cO(d)$ for both $g(\bx,k)$ and $h(\bx,\ell)$, if
$K = \sqrt{L}$, the time complexity for inference is reduced to
$\cO(b\times \sqrt{L}\times d)$ from $\cO(L\times d)$, which is the time
complexity of inference for a vanilla linear OVR model. Later in
Section~\ref{sec:xrlinear}, we will show that this time complexity can
be further reduced to $\cO(b \times \log L \times d)$ by a recursive approach.

It is worthwhile to mention that this framework also alleviates the data
sparsity issue that is a major issue in \xmc problems. By using clustering for semantic label indexing,
tail labels (i.e., labels with fewer positive instances) are clustered with
head labels (i.e., labels with more positive instances).
This allows the information from head labels to be ``transferred'' to tail
labels using our approach.
As a result, most indices (or clusters) contain more positive instances. For example,
the right panel of Figure~\ref{fig:label-distr-wiki}
shows the number of positive instances for each index/cluster after semantic label
indexing with $K=8,192$ clusters for an \xmc dataset with $L=500,000$ labels.
We can see that $99.4\%$ of the clusters contain more than $100$ training instances.

\begin{figure}[!t]
  \centering
  \includegraphics[width=0.95\linewidth]{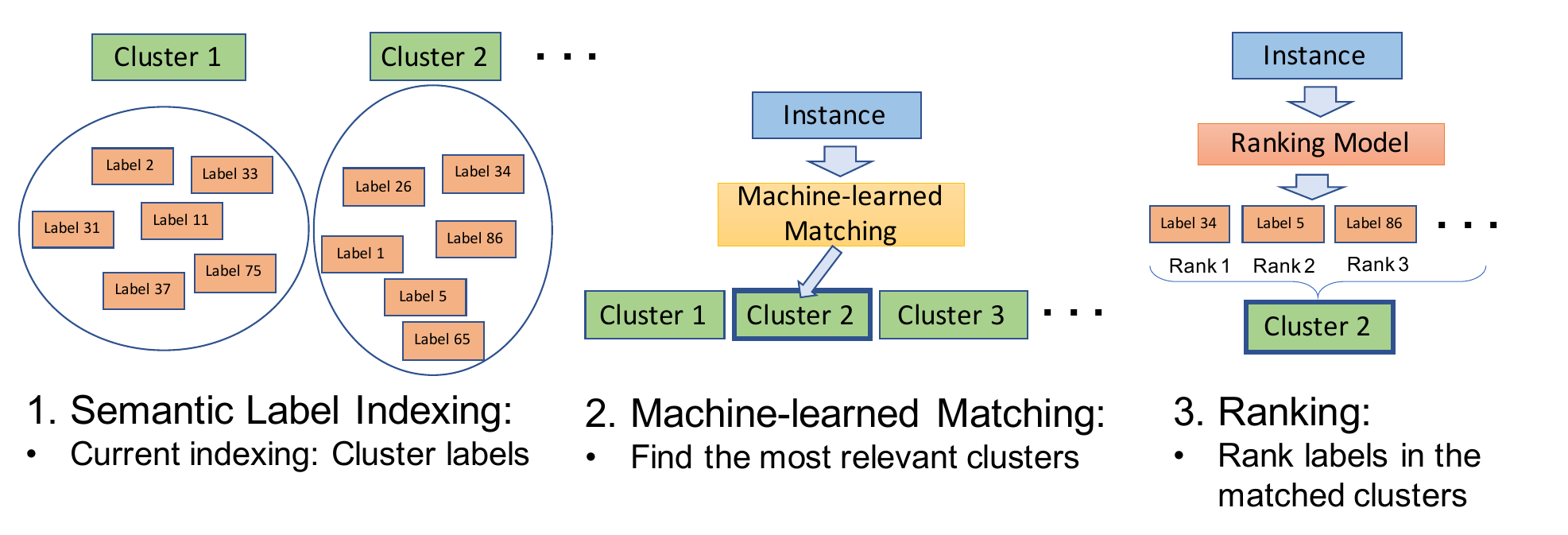}
  \caption{Illustration of the three-stage \pecos framework for \xmc.}
  \label{fig:pecos-framework}
\end{figure}

We now discuss each of the semantic indexing, matching and ranking phases in
greater detail.
\subsection{Semantic Label Indexing}
\label{sec:indexing}

Inducing label clustering with semantic meaning brings several advantages to
our framework.  The number of clusters $K$ is typically set to be much smaller
than the original label space $L$.  Our machine learned matcher $g(\bx,k)$
then needs to map the input to a cluster, which itself is an induced \xmc sub-problem
where the output space is of size $K$. This significantly reduces
computational cost and mitigates the data sparsity issue illustrated in
Figure~\ref{fig:label-distr-wiki}.  Furthermore, label clustering also
plays a crucial role in the learning of the ranker $h(\xb,\ell)$.  For example, only labels
within a cluster are used to construct ``hard'' negative instances for training the
ranker (more details are in Section~\ref{sec:ranking}).  During inference, ranking is
only performed for labels within the top-$b$ clusters predicted by our machine
learned matcher $g(\bx,k)$.
In some \xmc applications, labels may come with some meta information, such as taxonomy or
category information which can be used to naturally form a semantic label
clustering. However, when such information is not
explicitly available, we need to think about how to effectively perform semantic label
indexing.

There are two key components to achieve a good semantic label indexing:
{label representations} and {indexing/clustering algorithms}.

\subsubsection{Label Representations}
\label{sec:label-representations}

In general, label representations or label embeddings should encode the
semantic information such that two labels with high semantic similarity
have a high chance to be grouped together. We use $\cbr{\bz_{\ell}:\ell\in\cY}$ to denote the
label representations. If meta information is available for
labels, we can construct label representations directly from that information. For example, if
labels come with meaningful text descriptions such as the category information for Wiki
pages, we can use either traditional approaches such as term frequency-inverse document
frequency (\tfidf) or recent deep-learning based text embedding
approaches such as Word2Vec~\citep{mikolov2013distributed}, \elmo~\citep{peters2018deep} to
form label representations. If labels come with a graph structure such as
co-purchase graphs among items or friendship among users, one can consider
forming graph Laplacians~\citep{smola2013graph} or graph convolution neural networks~\citep{wu2019comprehensive} to
obtain label representations. Here, we present a few alternative ways to represent
labels when such meta information is not available.

{\bf Label Representation via Positive Instance Indices (\pii).}
\pii is a simple approach to represent each label by the membership of
its instances:
\[
  \bz_{\ell}^{\text{\pii}} := \frac{\bybar_{\ell}}{\norm{\bybar_{\ell}}},\quad \ell \in \cY,
\]
where $\bybar_{\ell} \in \cbr{0,1}^n$ is the $\ell$-th column of the instance-to-label
matrix $Y$.

{\bf Label Representation via Positive Instance Feature Aggregation (\pifa).}
In \pifa, each label is represented by aggregating feature vectors from
positive instances:
\begin{align*}
  \bz_{\ell}^{\text{\pifa}} &= \frac{\bv_{\ell}}{\norm{\bv_{\ell}}},\text{
  where } \ \bv_{\ell} = \sum_{i=1}^{L} Y_{i\ell}\bx_i = \rbr{X^\top Y}_{\ell}, \quad \ell \in \cY,
\end{align*}
where $\bx_i$ is the feature representation of the $i$-th training instance,
and $X=\sbr{\bx_1,\ldots,\bx_i,\ldots,\bx_n}^\top$ is the training instance
matrix. Note that the dimension of \pifa representations is $d$, which is
different from the dimension of \pii representations $n$.

{\bf Label Representation via Label Features in addition to \pifa (\pifa + \lf).}
If a label feature matrix
$\Ztil=[\bztil_1,\ldots,\bztil_{\ell},\ldots,\bztil_{L}]^\top \in \RR^{d \times L}$
is given and $\bztil_{\ell}$ and $\bx_i$ are in the same domain, we can
consider a weighted combination of $\bztil_{\ell}$ and \pifa representation as follows:
\[
  \bz_{\ell}^{\text{\pifa + \lf}} = (1-\alpha_{\ell})\bztil_{\ell} + \alpha_{\ell}
  \bz_{\ell}^{\text{\pifa}} = \rbr{(1-\alpha_{\ell})\Ztil + \alpha_{\ell} \rbr{X^\top Y}}_{\ell}.
\]

{\bf Label Representation via Graph Spectrum (\spectral).} In \spectral, we
consider the instance-to-label matrix $Y$ as a bi-partite graph between
instances and labels. We can then follow the co-clustering algorithm described
in  \citet{dhillon2001coclustering} to obtain spectral representations for
labels. In particular, we first form a normalized label matrix $\Ytil$ as follows
\[
  \Ytil = D_1^{-1/2} Y D_2^{-1/2},
\]
where $D_1 \in \RR^{n\times n} $, and $D_2\in \RR^{L\times L}$ are degree diagonal
matrices such that $(D_1)_{ii} = \sum_{\ell} Y_{i\ell}$ and
$(D_2)_{\ell\ell} = \sum_{i} Y_{i\ell}$. Next, let
$\cbr{(\bu_t, \bv_t): \bu_t \in \RR^{n},\ \bv_t \in \RR^{L},\ t=2,\ldots,k+1}$
be the singular vector pairs (left and right) corresponding to the 2nd,$\ldots$, $k+1$-st
largest singular values of $\Ytil$. Following \citet[Eq.
12]{dhillon2001coclustering}, we can construct a $k$-dimensional label representation matrix
as follows:
\begin{align*}
    \bz_{\ell}^{\text{\spectral}} &= \text{the $\ell$-th row of $Z$},\quad Z=D_2^{-1/2}[\bv_2,\ldots,\bv_{k+1}]^\top \in \RR^{L\times k}.
\end{align*}

\begin{figure}[t!]
  \centering
  \includegraphics[width=0.8\textwidth]{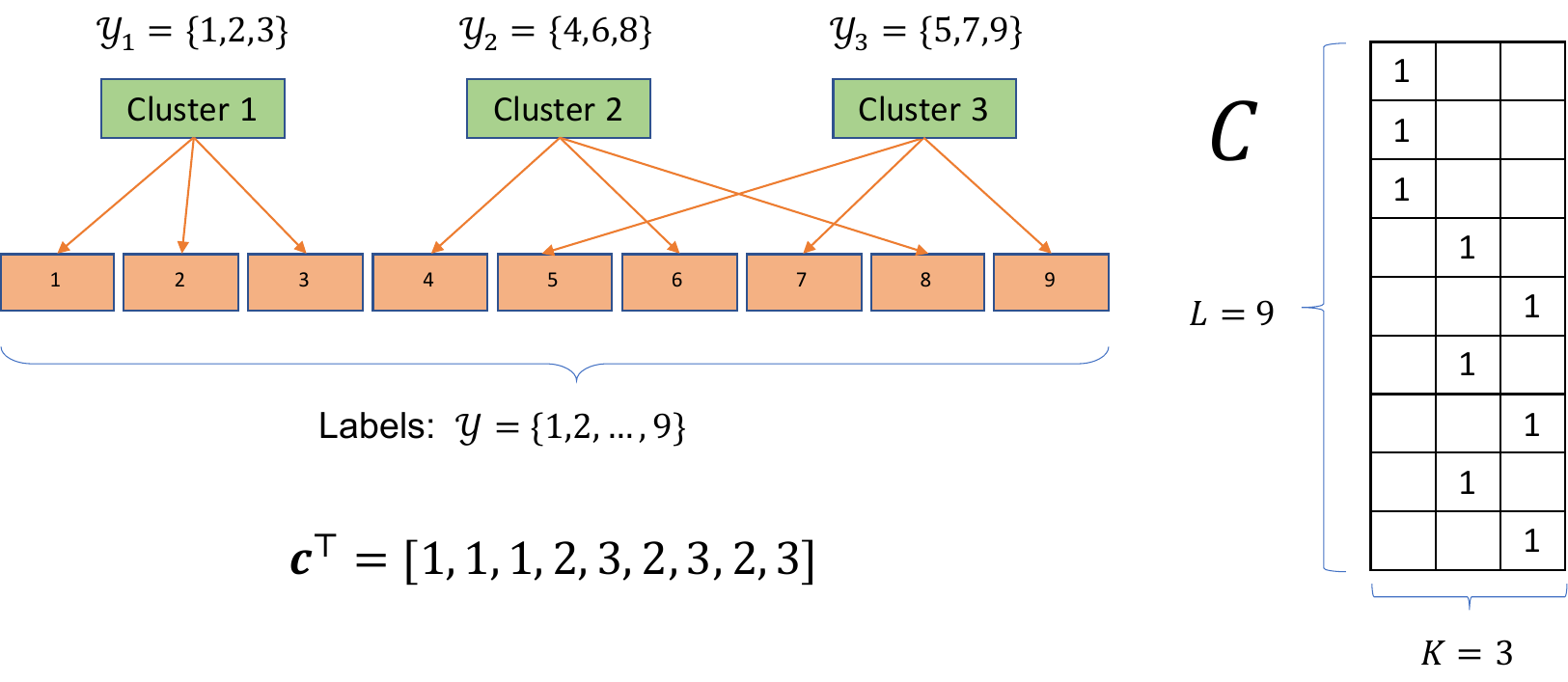}
  %\vspace{-1em}
  \caption{Illustration of Label Indexing/Clustering. $\bc \in \cbr{1,2,3}^9$ denotes
  the clustering vector with $c_{\ell}$ being the index of the cluster
  containing the $\ell$-th label, while $C\in\cbr{0,1}^{9\times 3}$ denotes
  the cluster indicator matrix (with same information as $\bc$).}
  \label{fig:indexing-matrix}
\end{figure}

See Figure~\ref{fig:indexing-matrix} for an illustration of the indexing
matrix with $L=9$ labels and $K=3$ clusters.

\subsubsection{Semantic Indexing Through Clustering}
\label{sec:clustering-algos}
Once the label representations $\cbr{\bz_{\ell}:\ell \in \cY}$ are decided, we can
get a semantic indexing  scheme using an appropriate clustering. Let
$\cC=\cbr{1,\ldots,K}$ denote the set of $K$ label clusters. The purpose of
our clustering algorithm is to learn a label-to-cluster
assignment: $\bc\in \cC^{L}$, where $c_{\ell}$ denotes the index of the cluster
containing the label $\ell$. Equivalently, the clustering assignment can also
be represented by the indexing matrix $C\in\cbr{0,1}^{L\times K}$ as follows:
\begin{align}
  C_{lk} &=
  \begin{cases}
  1, & \text{ if } k = c_{\ell}, \\
  0, & \text{ otherwise.}
  \end{cases}
  \label{eq:indexing-matrix}
\end{align}

Below we give the objective used in two popular $K$-Means~\citep{duda2012pattern} and
Spherical $K$-Means~\citep{dhillon2001concept} clustering algorithms:
\begin{align}
  \bc^{\text{K-MEANS}} = &\arg\min_{\bc \in \cC^L}\quad \sum_{k \in \cC} \sum_{\ell : c_{\ell} = k}
  \norm{\bz_{\ell} - \bmu_k}^2,\quad
  \text{where}\ \bmu_k := \frac{\sum_{\ell : c_{\ell} = k}\bz_{\ell}}{\abs{\cbr{l: c_{\ell} = k}}} ,  \label{eq:kmeans} \\
  \bc^{\text{SK-MEANS}} = &\arg\max_{\bc \in \cC^L}\quad \sum_{k \in \cC} \sum_{\ell : c_{\ell} = k}
  \frac{\bz_{\ell}^\top \bmu_k}{\norm{\bz_{\ell}}\norm{\bmu_k}},\quad
  \text{where}\ \bmu_k := \frac{\sum_{l: c_{\ell} = k} \bz_{\ell}}{\norm{\sum_{\ell : c_{\ell} = k} \bz_{\ell}}}.
 \label{eq:skmeans}
\end{align}

Standard $K$-Means/Spherical $K$-Means algorithms have
$\cO(\nnz(Z)\times K\times \text{\# iterations})$ computation complexity, where $Z$
is the label representation matrix. This can still be
very time consuming if $K$ is also large (say $K=10^4$). In \pecos, we provide
an implementation which utilizes a recursive $B$-ary partitioning approach to further
improve the efficiency of label clustering, see
Algorithm~\ref{alg:bisection-kmeans} The high level idea is to apply
$B$-ary  partitioning on the label set by either $B$-Means or spherical $B$-Means
recursively. $B$ is usually chosen as a small constant such as $2$ or $16$.
An illustration with $B=2$ is given in
Figure~\ref{fig:bisection-kmeans}. The time complexity of
Algorithm~\ref{alg:bisection-kmeans} is
$\cO(\nnz(Z)\times\log_{B} K \times \text{\# iterations})$, which is much lower
than directly clustering into $K$ clusters, which would have time complexity that is linear in $K$.

\begin{tabular}{@{}l@{}r}
    \hspace{-2em}
		\begin{minipage}[c]{0.48\textwidth}
			\begin{algorithm}[H]
				\captionof{algorithm}{Clustering with $B$-ary partitions}
				\label{alg:bisection-kmeans}
				\begin{compactitem}
				\item[] \textbf{Input:}
					\begin{compactitem}
					\item $\cY$: label indices and $\cbr{\bz_{\ell}: \ell \in \cY}$ representations,
          \item $K = B^{D-1}$: number of clusters.
					\end{compactitem}
				\item[] \textbf{Output:} indexing matrix: $C \in \cbr{0,1}^{L\times K}$
        \item $\cY^{(1)}_1 \leftarrow \cY$
				\item For $t=1,\ldots,D$
					\begin{compactitem}
					\item For $k=1,\ldots,B^{t-1}$
						\begin{compactitem}
            \item perform either $B$-Means or  Spherical $B$-Means to partition $\cY^{(t)}_{k}$ into $B$ clusters
            \item[] $\cbr{\cY^{(t+1)}_{B(k-1) + j}: j=1,\ldots,B}$
						%\item[] by either $B$-Means or Spherical $B$-Means
            %\item assign $\cY^{(t+1)}_{2k-1} \leftarrow \cA$
            %\item assign $\cY^{(t+1)}_{2k} \leftarrow \cB$
						\end{compactitem}
					\end{compactitem}
				\item construct $C$ with
					$C_{lk} = \begin{cases}
            1 & \text{ if } \ell \in \cY^{(D+1)}_k, \\
						0 & \text{ otherwise.}
					\end{cases}$
				\end{compactitem}
			\end{algorithm}
		\end{minipage}
		\begin{minipage}[c]{0.55\textwidth}
			\includegraphics[width=1\textwidth]{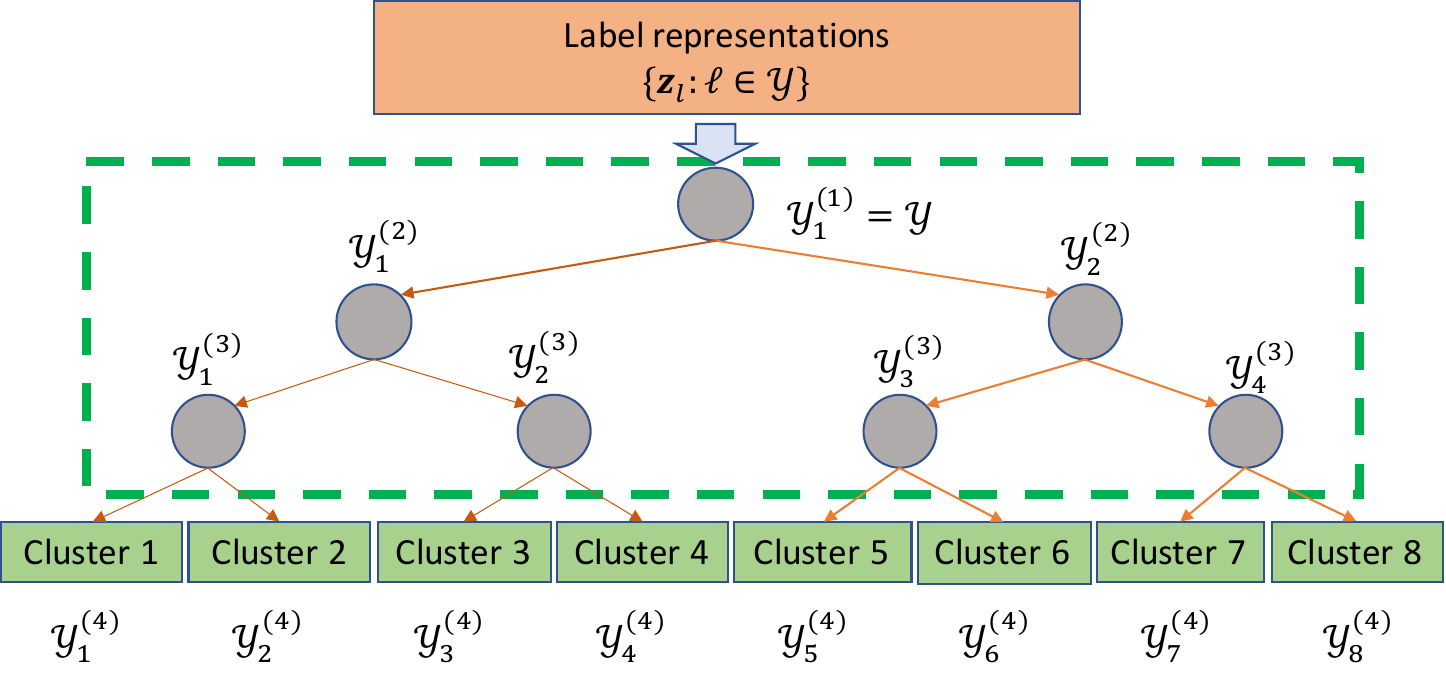}
			\captionof{figure}{Illustration of label clustering with recursive $B$-ary
      partitions with $B=2$. }
			\label{fig:bisection-kmeans}
		\end{minipage}
	\end{tabular}
\subsubsection{Other Indexing Methods}
\label{sec:other-indexing}
In this paper, we mainly focus on {\em label clustering} as the
algorithmic choice for the semantic label indexing phase. There are other
strategies which might be promising alternatives for \pecos, which we leave
as direction for future exploration. When using typical clustering algorithms
like $K$-Means each label is assigned to exactly one label cluster. However, labels in real
world applications might have more than one semantic meaning. For example,
``apple'' could be either a fruit or a brand. Thus, one interesting direction
to explore is to adopt overlapping clustering algorithms such as
\citet{whang2019non} for semantic label indexing.
Recently, \citet{liu2021label} propose to find overlapped clusters by
jointly optimizing cluster assignments and model parameters of partition-based \xmc models,
which is complementary to our \pecos framework.

In addition, we can use
various approximate nearest neighbor (ANN) search~\citep{li2019approximate} schemes as
the basis of semantic indexing.  There is a successful attempt by \citet{jain2019slice}
to apply a state-of-the-art approx ANN search algorithm called hierarchical
navigable small world (HNSW) graphs~\citep{malkov2020hnsw} for the \xmc problem when the
feature vectors are low-dimensional dense embeddings. The ability of variants
of the HNSW data structure such as Rand-NSG~\citep{subramanya2019rand} to
quickly get a small match set for any given input would be very suitable for the
three-stage design framework of \pecos.

\subsection{Machine Learned Matching}
\label{sec:matching}
The matching stage in \pecos is crucial since the final ranking stage is
restricted to the labels returned by the matcher; hence if the matcher fails
to identify the candidate labels accurately, performance can greatly suffer.
In general, the input features $\bx_i$ and label
representations $\bz_{\ell}$ could be in different domains and have different
dimensionalities. Thus, in the machine
learned matching stage,  we need to learn a general matcher function
$g(\bx, k)$ which finds the relevance between a given instance $\bx$ and the
$k$-th label cluster. This matching scoring function can then be used to
obtain the top-$b$ label clusters:
\[
  g_b(\bx) = \mathop{\arg\max}_{\cS\subset\cC:\abs{\cS}=b} \sum_{k \in \cS} g(\bx, k).
\]
%which can be used to shortlist label candidates in the inference. \isd{expand on this, write it in math.}

Given a semantic label indexing denoted by the clustering matrix $C\in\cbr{0,1}^{L\times K}$, the
original \xmc problem with the output space $\cY$ morphs to an \xmc sub-problem
with a much smaller output space $\cC$ of size $K$. In particular, we can
transform the original training dataset $\cbr{(\bx_i, \by_i):i=1,\ldots,n}$ to
a new dataset $\cbr{(\bx_i, \bbm_i): i=1,\ldots,n}$, where
$\bbm_i \in \cbr{0,1}^K = \text{binarize}\rbr{C^\top \by_i}$ denotes the ground truth input-to-cluster
assignment for the $i$-th
training instance. Similar to the instance-to-label matrix $Y$, we can stack the ground
truth $\cbr{\bmm_i}$ into the ground truth input-to-cluster assignment matrix
$M=[\bbm_1,\ldots,\bbm_i,\ldots,\bbm_n]^\top \in \cbr{0,1}^{n\times K}$. For
any given indexing matrix $C$, the ground truth $M$ can be obtained by
\[
  M = \text{binarize}(\Mtil),\text{ where }\ \Mtil = Y C,
\]
where $M_{ik} = \Ind{\sum_{\ell\in\cY} Y_{i\ell} C_{lk} > 0 }$. Note that if we
wanted a weighted input-to-cluster matrix, we could work with $\Mtil$ instead.

Thus, the machine learned matcher reduces to an \xmc problem with a smaller output space of
size $K$. Hence, we can apply any existing multi-label classifier which can handle $K$
labels. For example, if $K$ is not very large, we can consider the
aforementioned vanilla one-versus-rest (OVR) approach to learn the matcher
$g(\bx, k)$. If $K$ is still too large, we can recursively
apply the three-stage \pecos framework to learn the matcher. We will
give an example of this recursive \pecos approach, called \xrlinear, in
Section~\ref{sec:xrlinear}.

%Note that the matcher $g(\bx, k)$ plays a key role in producing a shortlist of label
%candidates to be considered for the final ranking stage.
Note that if the cluster of a relevant label is not correctly predicted by the matcher,
this relevant label does not have a chance to be surfaced by our ranker at all.
Thus, in Section~\ref{sec:deep-matcher},
we consider using more advanced deep learning based approaches to learn the matcher,
especially when the input instances are in text form.

\subsection{Ranking}
\label{sec:ranking}
The goal of the ranker $h(\bx, \ell)$ is to model the relevance
between the input $\bx$ and the shortlisted labels obtained from the relevant label clusters
identified by our matcher $g_b(\bx)$. Informally, the shortlist of candidate
labels is the set of label clusters. Given a label-to-cluster assignment
vector $\bc\in\cbr{1,\ldots,K}^L$, where the $k$-th cluster is given by $\cY_k=\cbr{\ell \in \cY: c_{\ell}=k}$, the
``shortlisting'' operation for an input $\bx$ can be formally described by
$s(\bmbar|\bc)$ as follows:
\begin{align}
  s(\bmbar|\bc) &= \bigcup\limits_{k:\mbar_k \neq 0} \cY_k, \label{eq:shortlist}
\end{align}
where $\bmbar\in \cbr{0,1}^K$ is the cluster indicator vector for the input $\bx$.
Here $\mbar_k =1$ denotes that the $k$-th cluster is considered relevant to the
input $\bx$. In general, for the $i$-th input $\bx_i$, this indicator vector
$\bmbar_i$ can come from either the ground truth input-to-cluster assignment
$\cbr{\bmm_i}$ defined in Section~\ref{sec:matching} or the relevant clusters
predicted by our machine learned matcher $\cbr{\bmhat_i}$, where the details
are provided in Section~\ref{sec:ranking-negative-sampling}. Note that, given
the clustering, $\bmm_i$ is an induced static assignment while the
choice of $\bmhat_i$ from the machine learned matcher depends on the
predictions made by the matcher. In particular, we
use $\bmhat$ to denote the indicator vector of label
clusters predicted by our matcher $g_b(\bx)$:
\[
  \mhat_k = \begin{cases}
    1 & \text{ if } k \in g_b(\bx), \\
    0 & \text{ otherwise. }
  \end{cases}
\]

The ideal ranker $h(\bx,\ell)$ for the given matcher $g_b(\bx)$ should satisfy the property:
\begin{align}
  h(\bx, \ell_1) > h(\bx, \ell_2) \Leftrightarrow \ell_1 \succ_{\bx} \ell_2\quad\forall
  \ell_1,\ell_2 \in s(\bmhat|\bc),
  \label{eq:ideal-ranker}
\end{align}
where $\ell_1 \succ_{\bx} \ell_2$ denotes that label $\ell_1$ is more relevant than
label $\ell_2$ for the input $\bx$ in the ordering of the ground truth.

In general, one can choose any ML ranking model as the ranker. Common choices include
linear models, gradient boosting decision trees (GBDT) and neural nets. Choices
of the loss function include point-wise, pair-wise, and list-wise ranking
losses. The modeling of \pecos allows for easy inclusion of
various rankers.

\subsubsection{Hard Negative Sampling for Ranker Training in \pecos}
\label{sec:ranking-negative-sampling}
One of the key components in learning a ranking model is to identify the sets of
positive (relevant) and negative (irrelevant) labels for each instance.
Unlike most standard ranking problems where positive and negative labels for
each instance are explicitly provided in the training dataset, for each instance
in an \xmc problem, we are usually provided a small number of {\em explicit}
relevant labels and abundant {\em implicit} irrelevant labels. Obviously,
including all implicit irrelevant labels as negative labels to train a ranker
is not feasible as it greatly increases training time with little increase in
accuracy. Hence, we propose to include only {\em hard
negatives}, by restricting them to be irrelevant labels from relevant label
clusters for each instance. In
particular, for a given instance $\bx_i$, letting $\bmbar_i$ be the indicator vector
for the relevant clusters, we have
\begin{align}
  \text{positives}(\bx_i) &= \cbr{\ell \in s(\bmbar_i|\bc): Y_{i\ell} = 1},\notag \\
  \text{negatives}(\bx_i) &= \cbr{\ell \in s(\bmbar_i|\bc): Y_{i\ell} \neq 1},
  \label{eq:ranking-samples}
\end{align}
where $s(\bmbar_i|\bc)$ is as in \eqref{eq:shortlist}, and gives the shortlist
candidate set of labels for training instance $\bx_i$.

Depending on the choice of the indicator vectors $\cbr{\bmbar_i}$, we have
the following hard negative sampling schemes.

\textbf{Teacher Forcing Negatives~(TFN)}.
{\em Teacher forcing}~\citep{williams1989learning,lamb2016professor} is a known training strategy used in
recurrent neural networks (RNN), where the ground truth for earlier outputs
is fed back into RNN training to be conditioned on for the prediction of
later outputs. In our framework, we use {\em  teacher forcing negatives (TFN)}
to denote the hard negative sampling scheme where the ground-truth
input-to-cluster assignment for the input $\bx_i$ is used to identify hard
negative labels for the training of the ranker.  In particular, the
input-to-cluster assignment is chosen as follows:
\[
\bmbar_i \leftarrow \bmm_i,\ \forall i,
\]
where $\bmm_i$ is the ground-truth input-to-cluster assigned used to train our
matcher in Section~\ref{sec:matching}.
As discovered in \citet{bengio2015scheduled}, the teacher forcing scheme can
lead to a discrepancy between training and inference for recurrent models. In
particular, during inference, the unknown ground truth is replaced by the
prediction generated by the model itself. This discrepancy leads to sub-optimal
performance for the models trained with the teacher forcing strategy.
Similarly, this discrepancy also appears in the TFN sampling scheme for
inferring our hard negatives as $\bmm_i$ is independent of the performance of
our matcher.

\begin{figure}[t!]
  \centering
  \includegraphics[width=0.95\textwidth]{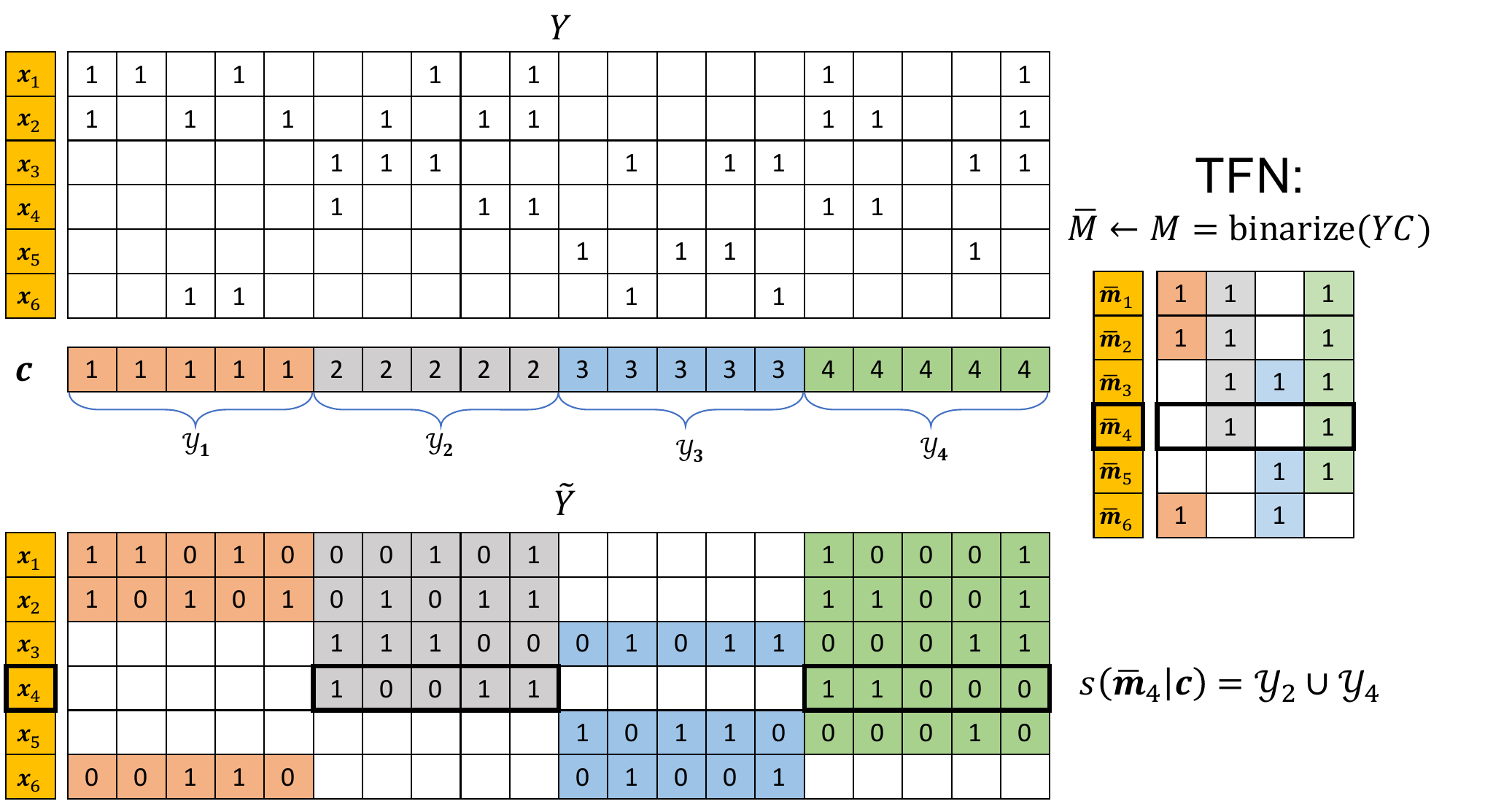}
  \caption{Illustration of Hard Negative Sampling for Ranker Training in \pecos.
In this toy \xmc example, we have $n=6$ instances, $L=20$ labels, and $K=4$ label
clusters. $Y$ shown on the top, denotes the ground-truth input-to-label matrix where
explicit positives are denoted by $1$. $\bc\in\cbr{1,\ldots,4}^{20}$ is the
label-to-cluster assignment vector for the label clustering
$\cbr{\cY_k:k=1,\ldots,4}$. We illustrate how to identify
the set of shortlisted labels given a input-to-cluster indicator vector
$\bmm_i$. (TFN chooses $\Mbar$ to be $M$:
$\Mbar\leftarrow M =[\cdots\bmm_i\cdots]^\top$ is the matrix obtained by
stacking $\cbr{\bmm_i}$).  Taking $\bx_4$ as an example, the positive labels
are $\by_4^{+}=\cbr{6,9,10,16,17}$. $\bmbar_4$ is an
example instance-to-cluster indicator where only the second and the fourth
clusters are chosen: $\bmm_4\leftarrow [0, 1, 0, 1]^\top$. As a result, only
labels from the shortlisted candidates $s(\bmbar_4|\bc)=\cY_2 \cup \cY_4=\cbr{6,7,8,9,10,16,17,18,19,20}$ are
considered in the training of the ranker for instance $\bx_4$. In particular, the negative
labels for $\bx_4$, denoted by cells with explicit zeros on the bottom, become
$s(\bmbar_4|\bc) \setminus \by_4^{+} = \cbr{7, 8, 18, 19, 20}$ instead of
$\cY \setminus \by_4^+$.
The choice of $\Mbar$ in above example follows teacher forcing negatives (TFN):
$\Mbar\leftarrow M$ (i.e., $\bmbar_i \leftarrow \bmm_i,\ \forall i$), which denotes
that the ground-truth input-to-cluster indicator
$\bmm_i=\text{binarize}(C^\top \by_i)$ is used to induce hard negatives for
each input. See Section~\ref{sec:ranking-negative-sampling} for more discussion about various
hard negative sampling schemes.
%    This example contains $n=6$ training instances, $L=20$ labels, and $K=4$ clusters.
%$Y$ is the instance-to-label matrix, and $\bc \in \cbr{1,2,3,4}^L$ is the label
%indexing vector. $\cY_{k}$ denotes the labels within the $k$-th cluster.
%$\Mbar\in\cbr{0,1}^{n\times K}$ denotes a given instance-to-cluster matrix.
%$s(\bmbar_i|\bc)$ defined in \eqref{eq:shortlist} denotes the shortlisted
%candidate labels by the given $\bc$ and $\bmbar_i$. In the ranking stage of
%\pecos, only irrelevant labels from the shortlisted candidate set induced by
%$\bmbar_i$ are included as negative labels for training. In the Teacher Forcing
%Negatives (TFN) scheme, we have $\bmbar_i =  \bmm_i = \text{binarize}(C^\top\by_i)$,
%while in the Matcher Aware Negatives (MAN) scheme, we have $\bmbar_i = \bmhat_i$,
%which is predicted cluster indicator by the matcher $g_b(\bx)$. In practice,
%we found that $\bmbar_i = \text{binarize}\rbr{\bmm_i + \bmhat_i}$ yields the
%best ranking performance. \isd{Need to add an example here. not clear what
%$\bmhat$ is in this figure.}
}
  \label{fig:ranking-negative-sampling}
\end{figure}

\textbf{Matcher Aware Negatives~(MAN)}.
An alternative strategy is to include matcher-aware hard negatives
for each training instance. In particular, for each input $\bx_i$, we can
use the instance-to-cluster indicator $\bmhat_i$ predicted by our matcher:
\[
  \bmbar_i \leftarrow \bmhat_i,\ \forall i.
\]
In practice, we observe that a union of TFN and MAN yields the best
performance:
\[
  \bmbar_i \leftarrow \text{binarize}\rbr{\bmm_i + \bmhat_i},\ \forall i.
\]
%i.e., using $\Mbar = \text{binarize}\rbr{M + \Mhat}$ to include hard negatives for each instance.
See Figure~\ref{fig:ranking-negative-sampling} for an illustration of how to identify
the set of shortlisted labels in order to train the OVR classifier for each
label, given an input-to-cluster indicator vector $\bmbar_i$.

\subsubsection{A One-Versus-Rest Linear Ranker}
\label{sec:simple-linear-ranker}
In general, we can use any ranker with a corresponding ranking loss function  in
\pecos. Here we present a simple one-versus-rest linear ranker with a
point-wise ranking loss. In particular, the linear ranker is parametrized by a
matrix $W = [\bw_1,\ldots,\bw_{\ell},\ldots,\bw_L]\in \RR^{d\times L}$ of
parameters as follows.
\[
  h(\bx, \ell) = \bw_{\ell}^\top\bx_i,\quad \ell \in \cY.
\]
Given an indexing vector $\bc$ and an instance-to-cluster matrix $\Mbar$, the
parameters $W$
can be obtained by solving the following optimization problem:
\begin{align}
  \min_{W}\quad &\sum_{i=1}^{n} \sum_{\ell \in s(\bmbar_i | \bc)} \cL(Y_{i\ell},
  \bw_{\ell}^\top\bx_i) + \frac{\lambda}{2} \sum_{\ell = 1} ^{L} \norm{\bw_{\ell}}^2,
  \label{eq:simple-linear-ranker-opt}
\end{align}
where $\cL(\cdot, \cdot)$ is a point-wise loss function such as
\begin{align}
  \cL_{\text{hinge}}(y, h) &= \max\cbr{0, 1 - \dot{y} h}, \notag\\
  \cL_{\text{squared-hinge}}(y, h) &= \max\cbr{0, 1 - \dot{y} h}^2, \notag\\
  \cL_{\text{logistic}}(y, h) &= \log\cbr{1 + \exp\rbr{- \dot{y} h}}, \notag
\end{align}
where $\dot{y} = 2 y - 1$, which maps $y$ from $\cbr{0,1}$ to $\cbr{-1, +1}$.
Due to the choice of point-wise loss, \eqref{eq:simple-linear-ranker-opt} can
be decomposed into $L$ independent binary classification problems as follows.
\begin{align}
  &\min_{W}\quad  \sum_{i=1}^{n} \sum_{\ell \in s(\bmbar_i | \bc)} \cL(Y_{i\ell},
  \bw_{\ell}^\top\bx_i) + \frac{\lambda}{2} \sum_{\ell = 1} ^{L}
  \norm{\bw_{\ell}}^2 \notag \\
 %\eqref{eq:simple-linear-ranker-opt} \equiv
  =&\quad \min_{W}\ \sum_{\ell \in \cY}
  \cbr{
    \sum_{i: \Mbar_{ic_{\ell}} \neq 0} \cL(Y_{i\ell}, \bw_{\ell}^\top\bx_i)
     + \frac{\lambda}{2} \norm{\bw_{\ell}}^2
   }\notag \\
   =&\quad  \cbr{
      \min_{\bw_{\ell}}\
      \sum_{i: \Mbar_{ic_{\ell}} \neq 0} \cL(Y_{i\ell}, \bw_{\ell}^\top\bx_i)
       + \frac{\lambda}{2} \norm{\bw_{\ell}}^2:\quad \ell \in \cY
    }
    \label{eq:independent-binary}
\end{align}
As a result, $\bw_{\ell}$ for label $\ell$ can be obtained, independent of
other labels, by any efficient solver for
the binary classification problem such as stochastic gradient descent (SGD) or
\liblinear~\citep{fan2008liblinear}. In the example of
Figure~\ref{fig:ranking-negative-sampling}, cells with colored
background in the bottom $Y$ matrix refer to shortlisted labels for each
input instance. For example, to train the OVR classification for label 2,
i.e., to compute $\bw_2$, we only include inputs corresponding to
the cells with colored background in the second column of the bottom $Y$
matrix: $\cbr{(\bx_1, Y_{12}=1), (\bx_2, Y_{22}=0), (\bx_6, Y_{62}=0)}$. We
use $\Ytil$ to denote the bottom sub-matrix containing only cells with colored
background.  Thus the training time for our approach is reduced to
$\text{nnz}(\Ytil)$ in
contrast to $n\times L$ which would be needed if all instances that are not positive
were used as negatives in the training procedure.
Furthermore, as each binary classifier can be
independently trained, we can apply various techniques to sparsify $\bw_{\ell}$
before we store it in memory, for example, by dropping zeros and small entries in
the computed $\bw_{\ell}$ parameters~\citep{babbar2017dismec,prabhu2018parabel}.
It is worth mentioning that, in practice, $L$ independent binary
classification problems can be computed in an embarrassingly parallel manner to
fully utilize the multi-core CPU design in modern hardware.

\subsection{Model Ensembling}
\label{sec:ensemble}
Model ensembling is a common and effective approach to further improve the
performance of machine learning models. There are two key components in model
ensembling: how to ensemble and what to ensemble. In terms of how to ensemble, many simple
strategies are considered and shown to be effective in many recent \xmc
approaches such as \citet{prabhu2018parabel,you2019attentionxml}. Options
include the averaging of the relevance score from individual models, the count of
being relevant from individual models, or the average candidate rank from
individual models. In terms of what models to ensemble, for \xmc, the existing
literature only uses limited options. Indeed, all of them consider
homogeneous models obtained by varying the random seed in some phases of the
training procedure, such as the random seed used to initialize the K-Means
clustering or the initial parameters~\citep{prabhu2018parabel}.

Due to its flexible three phase framework, \pecos offers a much more
sophisticated ensembling possibility. Thus, we propose to obtain an ensemble of
heterogeneous models obtained by various combinations of different
label representations, different label clusterings, different semantic
indexing schemes, different input feature representations, different machine learned matchers,
and different rankers. We have found that with the same number of models to
ensemble, an ensemble of heterogeneous models usually yields better
performance than ensembling homogeneous models. Due to the modularity and the flexibility of \pecos, this further
allows us to explore various combinations for each \xmc application.

\subsection{Inference}
\label{sec:inference}

In the inference phase of a \pecos \xmc model, we have a few options to obtain
the final relevance score $f(\bx, \ell)$. In general, it can be characterized as
follows.
\begin{align}
  f(\bx, \ell) =
  \begin{cases}
    \sigma(g(\bx, c_{\ell}), h(\bx, \ell))& \text{ if } \ell \in s(\bmhat | \bc), \\
    \inf\cbr{\sigma(g, h): g, h\in \RR}& \text{ otherwise,}
  \end{cases}
  \label{eq:inference-combiner}
\end{align}
where $\sigma(g, h)$ is a transform of the relevance scores from our
matcher $g(\bx, c_{\ell})$ and ranker $h(\bx, \ell)$. The time complexity of
inference is
\[
  \cO(\text{time to compute } g_b(\bx) + b\times \frac{L}{K} \times \text{time to compute } h(\bx, \ell)),
\]
where $b$ is the number of clusters predicted by our matcher (i.e., the so-called
beam size), and $L/K$ is the average number of labels in each label cluster $\cY_k$.

Here we discuss a few options for the transform function $\sigma(g, h)$. One option
is to only use the ranker score; using the matcher $g(\bx, \ell)$ only to
shortlist the label candidates in $s(\bmhat|\bc)$, i.e.,
\begin{equation}
	\sigma(g, h) = h.
	\label{eq:transform-noop}
\end{equation}
Another option is to consider the $\cL_p$-hinge transformation, i.e.,
\begin{equation}
  \sigma(g,h) = \exp\rbr{-\max(1 - g, 0)^p} \times \exp\rbr{-\max(1 - h, 0)^p},
	\label{eq:transform-lp-hinge}
\end{equation}
where $p=\{1,2,\ldots\}$.
The other option is to convert both $g$ and $h$ into probability values and
multiply the two probability values as the final score, e.g.,
\[
  \sigma(g, h) = \text{sigmoid}(g) \times \text{sigmoid}(h).
\]
In this case, one can give a probabilistic interpretation to the final value
$f(\bx, \ell)$ as follows:
\[
  f(\bx, \ell) = \text{Prob}\rbr{\text{$c_{\ell}$-th cluster} \mid \bx} \times \text{Prob}\rbr{\text{
  $\ell$-th label} \mid \bx, c_{\ell}},
\]
where $c_{\ell}$ is the label cluster containing label $\ell$.

\begin{algorithm}[t!]
  \caption{\xrlinear: a recursive realization of \pecos \xmc framework with
  simple linear rankers. }
  \label{alg:xrlinear}
  \begin{compactitem}
  \item[] \textbf{Input:}
    \begin{itemize}
      \item $X\in\RR^{n\times d}$: input feature matrix
      \item $Y\in\cbr{0,1}^{n\times L}$: input label matrix
      \item $\cbr{C^{(t)}: 1 \le t\le D}$:
        $ C^{(t)}\in \cbr{0,1}^{K_t\times K_{t-1}}$ indexing matrix at $t$-th
        layer with $K_{D} = L$ and $K_0 = 1$.
    \end{itemize}
  \item[] \textbf{Output:}
    \begin{itemize}
      \item $\cbr{h^{(t)}(\bx, k): 1 \le k \le K_t,\ 1\le t \le D}$:
        $h^{(t)}(\bx, k) = \bx^\top\bw_k^{(t)}$ ranker at $t$-th layer.
    \end{itemize}

  \item Form the training dataset for the \xmc problem at the $t$-th layer
    \begin{align*}
      X^{(t)} &\leftarrow X,\quad\quad\forall t=1,\ldots,D  \\
      Y^{(t)} &\leftarrow
      \begin{cases}
        Y & \text{ if } t = D, \\
        \text{binarize}\rbr{Y^{(t+1)}C^{(t+1)}} & \text{ if } t < D.
      \end{cases}
    \end{align*}
  \item Initialize a dummy \xmc model $f^{(0)}$:
    \begin{align*}
      &f^{(0)}(\bx, \ell) = 1\quad \forall \bx \in \RR^d, \ell \in
      \cbr{1,2,\ldots,K_1}, \\
      &f^{(0)}_b(\bx)  = \cbr{1,2,\ldots,K_1}.
    \end{align*}
  \item For $t=1,\ldots,D$
    \begin{itemize}
      \item Set the matcher to be the \xmc model obtained from the previous layer:
        \begin{align*}
          &g^{(t)}(\bx, k) = f^{(t-1)}(\bx, k),\quad k\in\cbr{1,2,\ldots,K_{t-1}} \\
          &g^{(t)}_b(\bx) = f^{(t-1)}_b(\bx)
        \end{align*}
      \item Select Negative Sampling Strategy for the ranker at $t$-th layer:
        \[
          \Mbar^{(t)}\leftarrow
        \begin{cases}
          M^{(t)} \equiv \text{binarize}\rbr{Y^{(t)} C^{(t)}} & \text{ Teacher Forcing Negatives (TFN)} \\
          \Mhat^{(t)} \equiv g^{(t)}_b(X)& \text{ Matcher Aware Negatives (MAN)} \\
          \text{binarize}\rbr{M^{(t)} + \Mhat^{(t)}} & \text{ TFN + MAN } \\
        \end{cases}
        \]
      \item Train the ranker $h^{(t)}(\bx, \ell)$ at the $t$-th layer with the
        parameter matrix $W^{(t)} \in \RR^{d\times K_t}$ where $\bw^{(t)}_{\ell}$
        is the $\ell$-th column obtained by solving:
        \[
          \bw_{\ell}^{(t)}=\arg\min_{\bw}\
        \sum_{i: \Mbar^{(t)}_{ic_{\ell}} \neq 0} \cL(Y^{(t)}_{i\ell}, \bw^\top\bx_i)
        + \frac{\lambda}{2} \norm{\bw}^2,\quad \ell=1,\ldots,K_t
        \]
        where $c_{\ell} = c^{(t)}_{\ell}  \in \cbr{1,\ldots,K_{t-1}}$ is the cluster
        index of the $\ell$-th label at the $t$-th layer.
      \item Obtain the \xmc model $f^{(t)}(\cdot)$ for the $t$-th layer, which
        will be used as the matcher for the $t+1$-st layer:
        \[
          f^{(t)}(\bx, \ell) =
          \begin{cases}
            \sigma^{(t)}\rbr{g^{(t)}(\bx, c_{\ell}^{(t)}), h^{(t)}(\bx, \ell)} & \text{ if } \ell
            \in s(\bmhat|\bc^{(t)}), \\
            -\infty & \text{otherwise},
          \end{cases}
        \]
        where $\bmhat \in \cbr{0,1}^{K_{t-1}}$ is induced by $g^{(t)}_b(\bx)$,
        and $\bc^{(t)}$ is the indexing vector corresponding to $C^{(t)}$.
    \end{itemize}
  \end{compactitem}
\end{algorithm}

\begin{figure}[t!]
  \centering
  \includegraphics[width=0.95\textwidth]{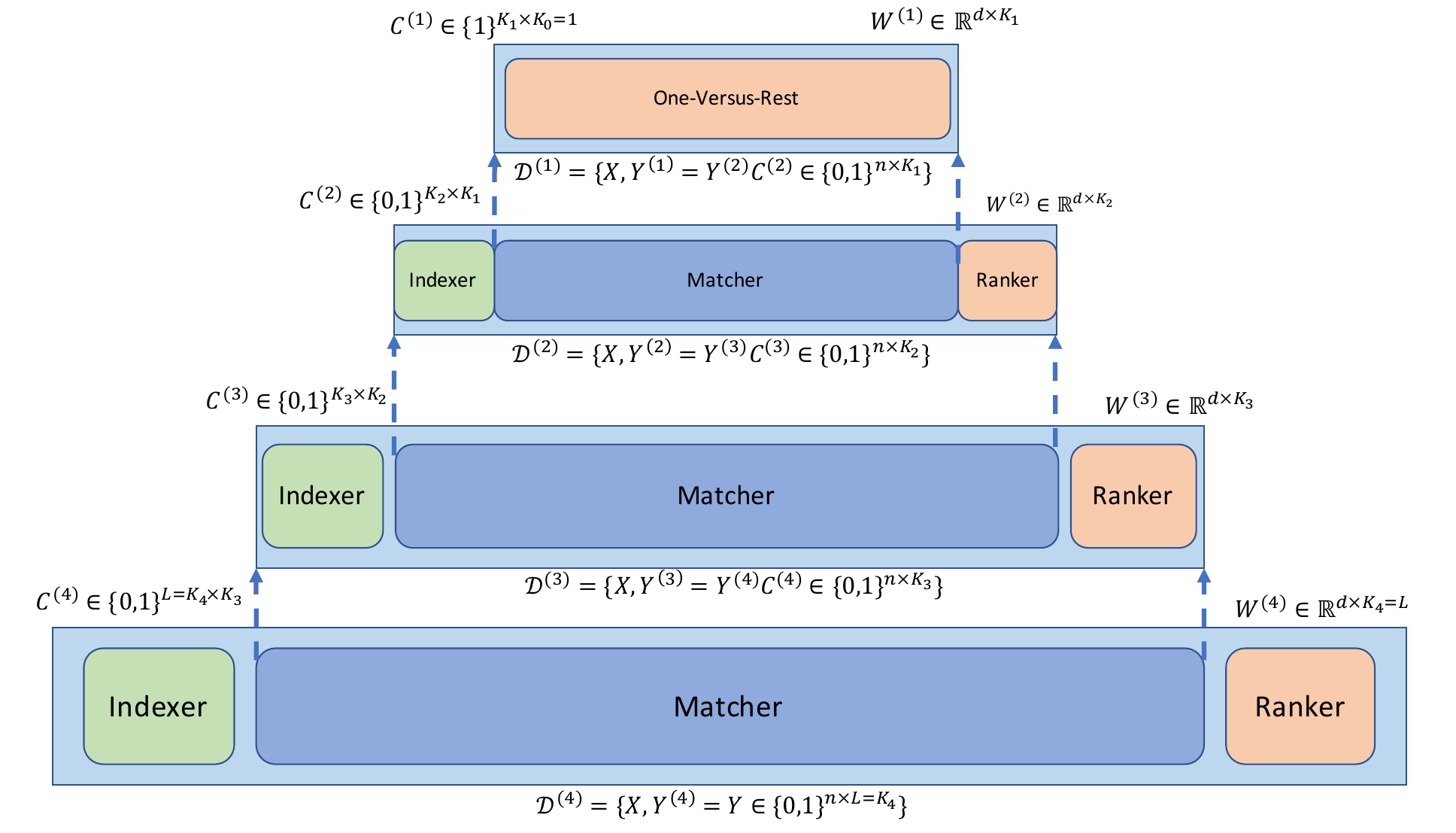}
  \caption{Illustration of \xrlinear.}
  \label{fig:xrlinear}
\end{figure}

\section{\xrlinear}
\label{sec:xrlinear}
In this section, we present \xrlinear, a recursive realization of our \pecos
framework proposed in Section~\ref{sec:pecos-framework}. In
particular, we exploit the property that the sub-problem handled by the
matcher is also an \xmc problem with a smaller output space of size $K$. Thus,
we can further apply the three stage \pecos framework recursively.

%An alternative format is a vector of pairs whose first elements are ordered feature indices and second elements are the corresponding non empty rows.
%Given an index-ordered sparse input, we alternatively find the common non-zero feature indices using binary search progressively, i.e., the common indices are found in order. We found the inference time of this format is similar to the hash table format.
Let $\cbr{X, Y}$ be the training matrices for the original \xmc problem with
$X\in\RR^{n\times d}$ and $Y\in\cbr{0,1}^{n\times L}$. Given an indexing
matrix $C\in\cbr{0,1}^{L\times K}$, the ranker $h(\bx, \ell)$ can be trained
on $\cbr{X, Y}$ with negatives induced by $M=\text{binarize}\rbr{YC}$ and/or
$\Mhat$, which is the predicted instance-to-cluster matrix by the matcher
$g_b(\bx)$ on the training feature matrix $X$. For the choice of ranker in
\xrlinear, we consider the simple linear ranker proposed in
Section~\ref{sec:simple-linear-ranker}.  On the other hand, the training data
to train the matcher $g(\bx, k)$ is $\cbr{X, M}$.  If $K$ is small enough, we
can apply an OVR ranker or classifier to obtain $g(\bx, k)$; otherwise, we can
treat $\cbr{X'=X, Y'=M}$ as a smaller \xmc problem and apply the \pecos
3-stage framework to learn the matcher. In particular, what we need is a
smaller indexing matrix $C' \in \cbr{0,1}^{L'\times K'}$, where $L'=K$ is the
size of the output space of the matcher.

In \xrlinear, we apply the above procedure recursively $D$ times. In
particular, let
\begin{align}
  \cbr{C^{t} \in \cbr{0,1}^{K_{t}\times K_{t-1}}: K_0=1,\ K_D=L,\ t = 1,\ldots,D}
\label{eq:Clist}
\end{align}
be a series of indexing matrices used for each of the $D$ \xmc sub-problems. When
$t=D$, it corresponds to the original \xmc problem on the given training
dataset $\cbr{X^{(D)}=X,Y^{(D)}=Y}$. When $t=D-1$, we construct the \xmc
sub-problem induced by the matcher with the training dataset
$\cbr{X, \text{binarize}\rbr{Y^{D}C^{D}}}$.  In general, the sub-problem for
the matcher at the $t$-th layer forms a full \xmc problem at the $(t-1)$-st layer.
When $t=1$, the output space of the \xmc sub-problem is small enough to be
solved directly by an OVR ranker.  In
Algorithm~\ref{alg:xrlinear}, we present detailed steps on how to apply the
three stage \pecos framework $D$ times in order to solve the original \xmc
problem. In Figure~\ref{fig:xrlinear}, we give an illustration by using a
toy example with $D=4$. It is worth mentioning two special cases of \xrlinear.
When $D=2$, \xrlinear is the same as the standard non-recursive three stage \pecos with an
OVR linear matcher and a linear ranker. On the other hand, when $D=1$,
\xrlinear is equivalent to vanilla linear OVR over all the labels.

\paragraph{\bf Model Sparsification.}
An \xrlinear model is composed of $D$ rankers $h^{(t)}(\bx, \ell)$ parametrized by
matrices $W^{t}\in \RR^{d\times K_t}$. As mentioned earlier in
Section~\ref{sec:xmc-formulation}, naively storing the entire dense
parameter matrices is not feasible. To overcome a prohibitive model size, we apply a common
strategy~\citep{babbar2017dismec,prabhu2018parabel} to sparsify $W^{(t)}$. In
particular, after the training process of each binary classification to obtain
$\bw^{(t)}_{\ell}$, we perform a (hard) thresholding operation to truncate parameters with
magnitude smaller than a user given value $\epsilon\ge 0$ to zero. We can
choose $\epsilon$ approximately so that the parameter matrices can be stored in the main memory. Model
sparsification is essential to avoid running out of memory when both the number
of input features $d$ and the number of labels $L$ are large. In addition to
hard thresholding, we also explored the option to include
$\norm{\bw}_1$ as the regularization and found that hard thresholding yields
slightly better performance than L1 regularization.

\paragraph{\bf Choice of Indexing Matrices.}
\xrlinear described in Algorithm~\ref{alg:xrlinear} is designed in a way to take
any series of indexing matrices of the form specified in~\eqref{eq:Clist},
which in fact can represent a family of hierarchical label clusterings.
This means that if the original label set $\cY$ comes with a hierarchy which
can be represented in a form as \eqref{eq:Clist}, this hierarchy can be
directly used within \xrlinear. On the other hand, when such a label hierarchy
is not available, we can still apply the semantic label indexing (clustering) approaches
described in Section~\ref{sec:indexing} to obtain a series of indexing
matrices. In particular, as a byproduct of
Algorithm~\ref{alg:bisection-kmeans} with $B$-ary partitions, when $K=B^{D-1}$ a
series of indexing matrices are naturally formed as follows.
$C^{(D)} = C \in \cbr{0,1}^{L\times K}$ and for $t < D$,
$C^{(t)} \in \cbr{0,1}^{B^{t}\times B^{t-1}}$ with
\begin{align*}
  \rbr{C^{(t)}}_{lk} & =
  \begin{cases}
    1 & \text{ if } \ceil{\frac{\ell}{B}} =k, \\
    0 & \text{ otherwise }
  \end{cases},\quad \forall\ 1 \le \ell \le B^{t+1},\ 1 \le k \le B^{t}.
\end{align*}
This is essentially balanced hierarchical label clustering. With the
choice of this hierarchical clustering, the size of the output spaces in the
$D$ \xmc problems are
\[
  K_{D}=L, K_{D-1}=B^{D-1}, K_{D-2}=B^{D-2}, \cdots, K_{1}=B^1,
\]
respectively.

\paragraph{\bf Choice of Negative Sampling Schemes and Transform functions}
\xrlinear in Algorithm~\ref{alg:xrlinear} is flexible to adopt a different choice
of negative sampling schemes and transform functions $\sigma^{(t)}(g,h)$ at
each layer. In general, the best choices for all the layers are data dependent
and can be obtained via a proper hyper-parameter tuning. After some
explorations, we observe that the following choice gives reasonably good
performance among all the datasets we tried: TFN with $\cL_3$-hinge
transformation~\eqref{eq:transform-lp-hinge} for the first $D-1$ layers, and
TFN + MAN with the shortlisting transform function~\eqref{eq:transform-noop} for the
$D$-th layer.

\begin{figure}[t!]
  \centering
  \includegraphics[width=0.75\textwidth]{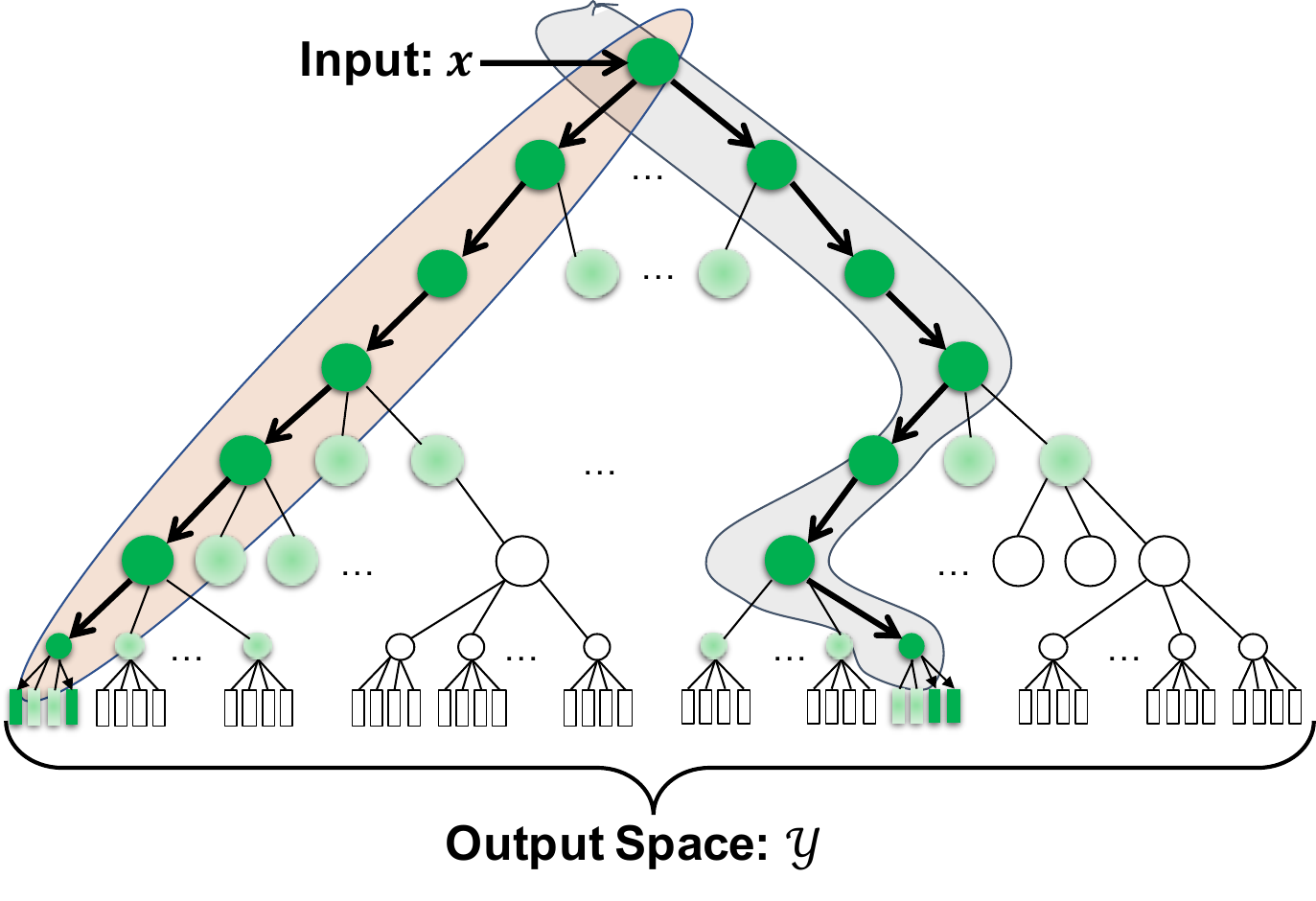}
  \caption{Illustration of the inference of \xrlinear using beam search with
  beam width $b=2$ to obtain $4$ relevant label predictions for the given
input $\bx$. The circular internal nodes denote label clusters at different
levels of the label hierarchy, while the rectangular leaf nodes denote labels
$\ell \in \cY$. We indicate in green color the label clusters which have been
traversed during the beam search. Finally, the labels found relevant for the
input $\bx$ are the green rectangular leaf nodes. }
  \label{fig:xrlinear-beam-search}
\end{figure}
\subsection{Efficient Inference for \xrlinear}
\label{sec:xrlinear-inference}
With the above choice of indexing matrices,Choice of Negative the inference for \xrlinear can be
made very efficient with beam search. Beam search is a heuristic search
algorithm to explore a directed graph (hierarchical label tree in our case) with
a limited memory requirement. In particular, beam search is a variant of
breadth-first search which only stores at most $b$ states at each level to
further expand, where $b$ is also called beam size. In
Figure~\ref{fig:xrlinear-beam-search}, we give an illustration of how beam
search works to perform inference in an \xrlinear model. Let $T_h$ be the time to
compute $ h^{(t)}(\bx, \ell)$, the
time complexity of inference via beam search with beam size $b$ becomes
\begin{align*}
  &\quad\sum_{t=1}^{D} \cO\rbr{\text{beam size} \times \max_{1\le k\le B^{t}}\ \abs{\cY^{(t)}_k}
  \times T_h} \\
  = &\quad\sum_{t=1}^{D} \cO\rbr{b \times \frac{K_t}{K_{t-1}} \times
  T_h}\quad\quad\quad\quad\quad\quad\cdots\text{due to the balanced partitions} \\
  = &\quad\cO\rbr{D \times b \times \max\rbr{B, \frac{L}{B^{D-1}}} \times T_h}
  \quad\cdots\text{due to the $B$-ary partitions}.
\end{align*}
We can see that if $D$ and $B$ are chosen such that $L/B^{D-1}$ is a small constant
(i.e., $D= \cO(\log_B L)$) such as $100$, the overall time complexity of the
inference for \xrlinear is
\[
  \cO\rbr{\log L \times b \times T_h},
\]
which is logarithmic in the size of the original output space.

\begin{figure}[t]
%\begin{center}
\includegraphics[width=3.3in]{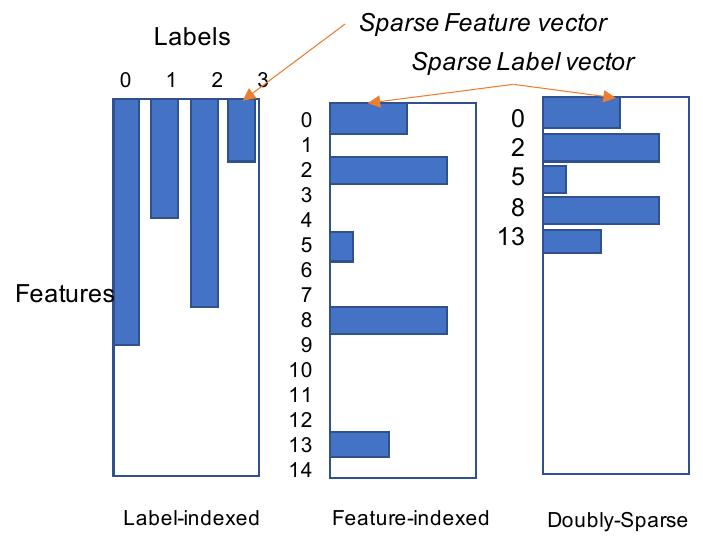}
\caption{Sparse Data Structures for the weight matrix
$W\in \RR^{d\times \abs{\cY}}$ in a label cluster, $\cY$. The memory
requirements from left to right are $\cO(|\cY| + \text{nnz}(W))$, $\cO(d + \text{nnz}(W))$ and
$\cO(\text{nnz}(W))$ respectively. %The computational complexities are
%$\cO(\abs{\cY}\times \text{nnz}(\bx) + \text{nnz}(W))$, $\cO( \text{nnz}(\bx) + \text{nnz}(W))$ and
%$\cO( \text{nnz}(\bx) + \text{nnz}(W))$ respectively.
}
\label{fig:struct}
%\end{center}
\end{figure}

\paragraph{\bf Efficient Ranking with Sparse Inputs.}
%The matching step retrieves a small subset of clusters after which we need to rank the labels in these clusters. As a ranking model, our goal is to
%model the relevance between the instance and the retrieved labels. Formally,
%given a label and an instance, we want to find a mapping $h(\bx, \ell)$ that
%maps the instance feature $\bx$ and the label $\ell$ into a score. In this
%paper, we mainly use the linear OVA approach, which treats the assignment of an individual label to an instance as an independent
%binary classification problem. The class label is positive if the instance
%belongs to the cluster; otherwise, it is negative. If the instance feature is
%in textual form, the input of the linear classifier can be the \tfidf feature. The output
%of the classifier is a probability that the instance belongs to the cluster.

Now we focus on how to efficiently rank the retrieved labels in real
time when the \xrlinear model weights and the input vectors are sparse. For a given input $\bx$, the score is defined
as $h(\bx, \ell) = \bw_{\ell}^\top \bx$, where $\bw_{\ell}$ is the weight vector for
the $\ell$-th label. For sparse input data, such as \tfidf features of text input,
$\bx$ is a sparse vector. By enforcing sparsity structure
on the weight vectors during training, a key computational step becomes the
multiplication of a sparse matrix and a sparse vector. However, many existing \xmc linear
classifiers, such as \parabel, are optimized for batch inference, i.e., the
average time is optimized for a large batch of testing data. In many
applications, we often need to do real-time inference, where the inputs
arrive one at a time.

Here we propose a data structure called doubly-sparse representation for the weight vectors along with an
algorithm to improve the speed of the real-time inference. Given a label
cluster $\cY \subset [L]$, we aggregate the weight vectors in this cluster and
form a weight matrix $W = [\bw_{\ell}]_{\ell\in\cY} \in \RR^{d\times |\cY|}$,
where $d$ is the number of features and $|\cY|$ is the number of labels in
this cluster. In Figure~\ref{fig:struct}, we illustrate several data structures to
store the weight matrix. In the label-indexed representation, we store
multiple (feature-index, value) pairs for each column of $W$. We
call each column vector as a sparse feature vector. Note that \parabel
\citep{prabhu2018parabel} implements the label-indexed representation. In
the feature-indexed representation, we store multiple (label-index,
value) pairs for each row of $W$. We call each row vector as a sparse label vector.
Each row of the weight matrix is recorded even if it is empty.  In the
doubly sparse representation, we store only the non-empty sparse label vectors
and the corresponding row indices.

Note that when the label cluster only contains a small set of
labels and the feature dimension is very large, the weight matrix will be very
sparse, i.e., $\text{nnz}(W) < d$, and the feature-indexed representation will consume
a lot of memory to store an empty vector for each feature. Therefore, to
achieve efficient real-time inference for a \xrlinear model, we propose to use
the doubly sparse representation, which is based on feature-indexed
representation but only stores non-empty label vectors. There are two data structures we can utilize to quickly find a given row index: 1) store the row indices of the non-empty label vectors in a sorted array and use binary search, 2) use a hash table to map the row indices to label vectors. For simplicity, we focus on using a hash table in this paper.

Given an input data, $\bx$, and a weight matrix $W$, our goal is to calculate
the scores for the labels, i.e., $W^\top \bx$. In the feature-indexed
representation, we can find the sparse label vector for the index of any non-zero feature
of the input $\bx$ in a constant time, therefore, the computational complexity
is $\cO(\text{nnz}(\bx) + \text{nnz}(W))$.  However, the memory requirement will be
$\cO(d+\text{nnz}(W))$ for each weight matrix. For some datasets, such as \wikil
($d \approx 500,000$ and $K\approx 5,000$), the total memory required can be
huge. In the label-indexed representation, $W^\top \bx$ consists of $\abs{\cY}$
inner products between two sparse vectors. As implemented in the \parabel code \citep{xmc_repo}, for
every inner-product, it first transforms a sparse weight vector to a dense
vector and then uses (sparse-matrix, dense-vector)
multiplication to calculate the inner product between the weight vector and the input vector. Therefore, the computational
complexity is $\cO(\abs{\cY}\times \text{nnz}(\bx) + \text{nnz}(W))$. To reduce both the
computational complexity and memory requirement, we use a doubly-sparse weight
matrix. We still use (label-index, value) pairs to store each non-empty row in
the weight matrix. We use a hash table
to  map the feature indices to the non-empty rows. Although the hashing step is slightly slower than direct memory access as in feature-indexed representation, it still has an amortized constant time.
Given a sparse input, we can get the corresponding rows from the non-zero
feature indices in constant time, therefore, the computational complexity
for calculating $W^\top \bx$ is only $\cO(\text{nnz}(\bx) + \text{nnz}(W))$ while
using $\cO(\text{nnz}(W))$ memory. We list the time and memory complexity of the overall inference in Table~\ref{tab:infer_complexity}.

\begin{table*}[!t]
    \centering
    \caption{Time and memory complexity of the inference. Here $b$: beam size.
    $\text{nnz}(\hat w)$: average number of non-zeros of the weight vectors.
  $D$: the depth of the tree ($D = \cO(\log L) $). }
    \label{tab:infer_complexity}
    \begin{tabular}{c|r|r|r|r|rrrr}
    Data Structure & Computational Complexity &     Memory Usage \\    \hline
    Label-indexed &   $b \times (\text{nnz}(\bx) + \text{nnz}(\hat{w})) \times (D + L/B^{D-1})$ & $ \text{nnz}(\hat{w}) \times L$   \\
    Feature-indexed  & $b \times \text{nnz}(\bx) \times D +  b \times \text{nnz}(\hat{w}) \times (D + L/B^{D-1})$ & $ \text{nnz}(\hat{w}) \times L + d \times B^{D}$  \\
    Doubly-sparse & $b \times \text{nnz}(\bx) \times D +  b \times \text{nnz}(\hat{w}) \times (D + L/B^{D-1})$ & $ \text{nnz}(\hat{w}) \times L $ \\
        \end{tabular}
\end{table*}

In summary, feature-indexed representation is the fastest but takes too much memory when $d$ is very large.
The doubly sparse representation is slightly slower than the feature-indexed representation
due to the hashing step but is much more memory-efficient.
Therefore, we recommend doubly sparse representation and
use it for the experimental results in Table~\ref{tab:exp-realtime-parabel} of Sec.~\ref{sec:exp-realtime}.
In \cite{etter2021accelerating}, we apply doubly sparse representation
(called {\it Masked Sparse Chunk Multiplication} in \cite{etter2021accelerating})
to different algorithms for sparse extreme multi-label ranking trees
and achieve faster inference than methods in \parabel~\citep{prabhu2018parabel}
and \napkinxc~\citep{jasinska2020probabilistic}.
%We will also use similar data structure for the matching step when it uses linear OVA approaches.

%\subsubsection{Inner Product between the instance feature and label features}
%During indexing, we have used some methods to construct label features, such as the indicator feature used in Homer and the sum of relevant instance features used in Parabel. If we use instance features to construct the label features, we can directly use the inner product between the label feature and the instance feature to score the relevance between a label and an instance. In particular, let $\xb_i$ and $\zb_j$ denote the instance feature and the label feature respectively. The score between them $h(\xb_i, \zb_j)$ is $\xb_i^\top \zb_j$.

\section{Deep Learned Matchers for Text Inputs}
\label{sec:deep-matcher}
In this section, we present {\em deep
learned} matchers for \xmc problems with {\em text inputs.} Note that unlike
\xrlinear which can handle general inputs where the features are in vector form, the
techniques discussed in this section only apply to \xmc problems with text
inputs. Let $\bt_i$ denote the text sequence associated with the $i$-th input.
Let $\bx=\bphi(\bt \mid \Theta)$
denote a vectorizer function which converts the input text sequence $\bt$
to a $d$-dimensional feature vector $\bx$, where $\Theta$ is the parameter
controlling the vectorizer. For example, a term frequency-inverse document
frequency (\tfidf) vectorizer, $\bphi_{\text{tfidf}}(\bt\mid\Theta_{\text{tfidf}})$
is parameterized by a vocabulary $\cV$ and the inverse document frequency of each term
$\cbr{\text{idf}(v): \forall v \in \cV}$. For an \xmc problem with text inputs,
\pecos with a \tfidf vectorizer works reasonably well in our experience.
However, the parameters $\Theta$ for traditional text vectorizers such as
\tfidf are obtained using only the text of the training set and are independent
from the supervision $Y$ provided in the training set.
%The same issue also arises in many natural language processing (NLP) tasks. \isd{please expand on this.}

\begin{figure}[t!]
  \centering
  \includegraphics[width=1\textwidth]{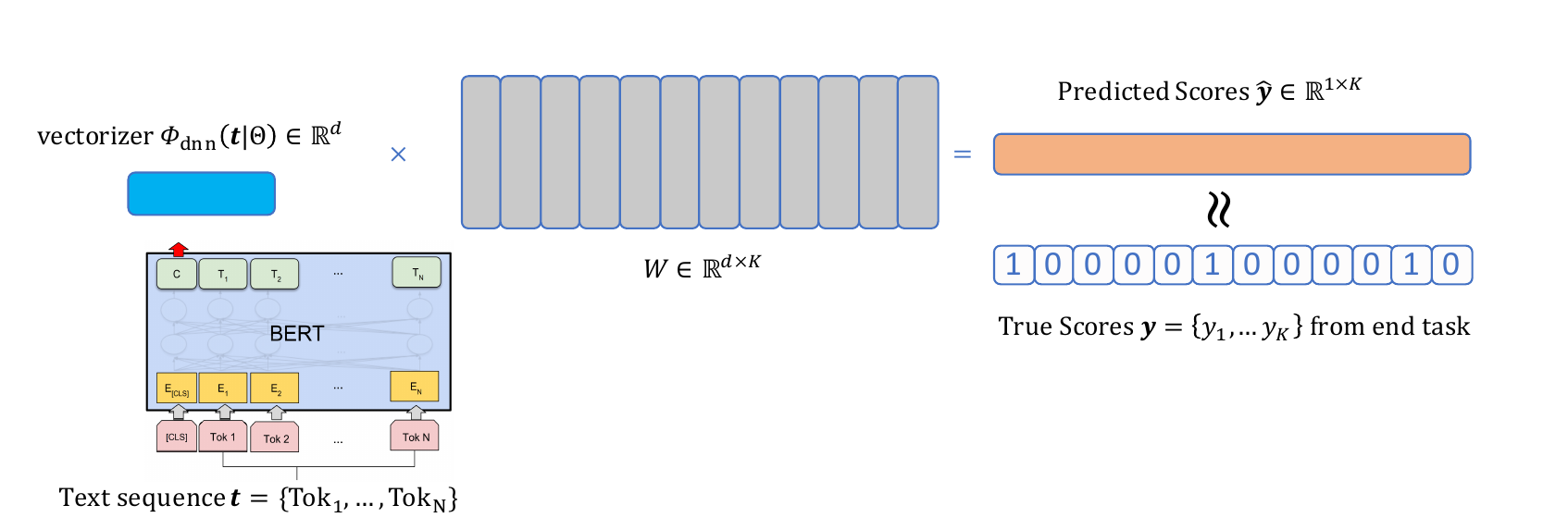}
  \caption{Illustration of fine-tuning a pre-trained Transformer model such as
  BERT~\citep{devlin2018bert} to a given end task.}
  \label{fig:finetune}
\end{figure}
Recently, deep pre-trained Transformers, e.g., BERT~\citep{devlin2018bert}
along with its many successors such as XLNet~\citep{yang2019xlnet} and
RoBERTa~\citep{liu2019roberta}, have led to state-of-the-art performance on
many NLP tasks, such as question answering, part-of-speech tagging, and
sentence classification with very few labels. Deep pretrained
Transformer models provide a trainable text vectorizer
that can be rapidly fine-tuned on many downstream NLP problems by adding a
task-specific lightweight linear layer on top of the Transformer models as
illustrated in Figure~\ref{fig:finetune}. In
particular, the text vectorizer from a given Transformer model can be
represented as $\bphi_{\text{dnn}}(\bt \mid \Theta)$, where $\Theta$
denotes the weights for the deep neural network architecture. Although the
pretrained $\Theta$ is usually obtained by learning a general language model
on a large text corpus, it can be fined-tuned on various downstream NLP tasks,
such as those in the GLUE benchmark~\citep{wang2018glue}.

We consider incorporating such a {\em trainable} deep text vectorizer
so we have a deep learned matcher:
\[
  g_{\text{dnn}}(\bt, k) = \bw_k^{\top} \bphi_{\text{dnn}}(\bt\mid\Theta).
\]
Note that the first argument of $g$ is the text sequence $\bt$ instead of the
feature vector $\bx$. Recall that the sub-problem to learn our matcher is also
an \xmc problem, where the training data is $\cbr{X, M}$. Note that when we
have millions of labels, using trainable deep vectorizers on the original
problem $\cbr{X, Y}$ would be prohibitive. If an OVR approach is used to learn
the matcher, we can solve the following fine-tuning problem to obtain the
parameters for our deep learned matcher:
\begin{align}
  \min_{\cbr{\bw_k}, \Theta}\quad \sum_{i=1}^{n} \sum_{k=1}^{K}
  \cL\rbr{M_{ik}, \bw_k^{\top} \bphi_{\text{dnn}}(\bt_i\mid\Theta)}, \label{eq:deep-matcher}
  %\cL\rbr{M_{ik}, g_{\text{dnn}}(\bt_i, k)}, \label{eq:deep-matcher}
\end{align}
where $\cL(\cdot,\cdot)$ is a loss function and $M = \text{binarize}(YC)$ is
the instance-to-cluster matrix. In particular, we use
$\cL_{\text{squared-hinge}}$ loss in our experiments as it has shown better performance in
existing \xmc work~\citep{yen2017ppdsparse,prabhu2018parabel}.
Due to the use of a deep text vectorizer $\bphi_{\text{dnn}}(\bt\mid\Theta)$
and having $\Theta$ as a trainable parameter in
\eqref{eq:deep-matcher}, we follow the providers of the pre-trained models and
use a variant of the Adam algorithm~\citep{kingma2014adam} to solve
\eqref{eq:deep-matcher}. Below we summarize our learnings in this
exploration.

\paragraph{\bf Choice of Deep Text Vectorizers.}
We consider three state-of-the-art pre-trained Transformer-large-cased models
(i.e., 24 layers with case-sensitive vocabulary) as our deep text vectorizers, namely
BERT~\citep{devlin2018bert}, XLNet~\citep{yang2019xlnet}, and
RoBERTa~\citep{liu2019roberta}.  In terms of training speed, BERT and RoBERTa are
similar while XLNet is nearly $1.8$ times slower.  In terms of performance on
\xmc tasks, we found RoBERTa and XLNet to be slightly better than BERT, but the
gap is not significant.

\paragraph{\bf Training Efficiency.}
The time and space complexity of the Transformer scales quadratically with the
input sequence length~\citep{vaswani2017attention}, i.e., $\cO(T^2)$, where
$T=\text{len}(\bt)$ is the number of tokenized sub-words in the instance $\bt$.
Using smaller $T$ reduces not only the GPU memory usage that supports using
larger batch size, but also increases the training speed.  For example,
BERT~\citep{devlin2018bert}
first pre-trains on inputs of sequence length $128$ for $90\%$ of the
optimization, and the remaining $10\%$ of optimization steps on inputs of
sequence length $512$.  Interestingly, we observe that
the model fine-tuned with sequence length $128$ v.s. sequence length $512$
does not differ significantly in downstream \xmc performance.
Thus, we fix the input sequence length to be $T=128$ for model fine-tuning,
which significantly speeds up the training time.  It would be interesting to
see if we can bootstrap training the Transformer models from shorter sequence
length and ramp up to larger sequence length (e.g., 32, 64, 128, 256), but we
leave that as future work.

\paragraph{Recursive Realization with Shared Encoder.}
Intuitively, the recursive realization with deep 
learned matchers would involve multiple deep learning text encoders.
However, in practice, sharing the text encoder across all the hierarchical 
layers would be a better choice. On one hand, the inference time is dominated by the 
evaluation of text embeddings and having multiple text encoders would greatly 
increase the inference latency. On the other hand, having a shared encoder means that 
the same neural network can be trained on multi-resolution label signals, 
which is proven to improve both training efficiency and model performance.  
More details can be found in \citet[Section 4]{zhang2021fast}.

\paragraph{\bf Further Utilization of Learned Deep Text Vectorizer.}
As a byproduct of our deep learned matcher, we have a powerful deep text
vectorizer $\bphi_{\text{dnn}}\rbr{\bt\mid\Theta_{\text{fnt}}}$, where
$\Theta_{\text{fnt}}$ denotes the fine-tuned parameters after solving
\eqref{eq:deep-matcher}. This vectorizer can be further utilized to further
improve the overall \xmc performance. First, we can concatenate it with a simple \tfidf
vectorizer to form the feature vector for our ranker. In particular, we can
have
\[
  \bx_i^\top = \sbr{
  \bphi_{\text{tfidf}}^\top\rbr{\bt_i\mid\Theta_{\text{tfidf}}}, \
    \bphi_{\text{dnn}}^\top\rbr{\bt_i\mid\Theta_{\text{fnt}}}
  }
\]
as the feature vector for the $i$-th instance to train the simple linear
ranker. We observe that such concatenation leads to the best overall
performance compared to the use of either \tfidf or Deep Text Vectorizer
individually. Second, as mentioned earlier in Section~\ref{sec:ensemble}, we
can use this deep learned text vectorizer to form a new set of feature vectors
and learn a new model based on it, which we can then ensemble with the model
based on \tfidf vectorizer.

\newpage
\section{Related Work}
\label{sec:related-work}

\subsection{Sparse Linear Models with Partitioning Techniques}
Conventional \xmc algorithms consider fixed input representations
such as sparse \tfidf features and leverage different partitioning
techniques or surrogate loss functions on the large label space to reduce complexity.
For example, sparse linear one-versus-reset~(OVR) methods such as DiSMEC~\citep{babbar2017dismec},
\proxml~\citep{babbar2019data}, PPDSparse~\citep{yen2016pd,yen2017ppdsparse} explore
parallelism to speed up the algorithm and reduce the model size by truncating model weights to encourage sparsity. 

The efficiency and scalability of OVR models can be further improved
by incorporating different partitioning techniques on the label spaces.  For
instance, \parabel~\citep{prabhu2018parabel} partitions the labels through a
balanced 2-means label tree using label features constructed from the instances.
Other approaches attempt to improve on \parabel,
for instance, \xtext~\citep{wydmuch2018no}, \bonsai~\citep{khandagale2020bonsai}, and
\napkinxc~\citep{jasinska2020probabilistic} relax two main constraints in
\parabel by: 1) allowing multi-way instead of binary partitions of the label set at each
intermediate node, and 2) removing strict balancing constraints on the partitions.
On the other hand, \slice~\citep{jain2019slice} and \annexml~\citep{tagami2017annexml}
partition the label spaces via graph-based approximate nearest neighbor (ANN) indices.
For a given instance,
relevant labels can be found quickly from nearest neighbors of the instance via the ANN graph.

\noindent \paragraph{\bf{Comparing \pecos with \parabel.}}
Concerning the three phase framework for \pecos,
we can interpret \parabel~\citep{prabhu2018parabel} as a special case of \xrlinear
with the following choices: \pifa label representation + Algorithm 1 (with $B=2$) + TFN sampling scheme.
There are three main differences between \xrlinear and \parabel.
First, \xrlinear generalizes \parabel with multi-way partitioning of the hierarchical label tree.
Second, \xrlinear incorporates various hard negative sampling schemes (e.g., MAN, TFN+MAN).
Finally, even if the model parameters are the same for \xrlinear and \parabel,
\xrlinear achieves significantly lower real-time inference latency
because of the doubly-sparse data structure described in Sections~\ref{sec:xrlinear-inference} and ~\ref{sec:exp-realtime}.  

\subsection{Neural Embedding-based Models}
Neural-based \xmc models employ various network architectures to learn semantic embeddings of the input text.
\xmlcnn~\citep{liu2017deep} employs one-dimensional CNN on the input sequence
and train the model with binary cross entropy loss without sampling, which is not scalable to large label spaces.
Shallow embedding-based methods aggregate word embeddings of a text input
followed by shallow MLP layers to obtain input embeddings,
which has smaller encoding latency for real-time inference.
Specifically, \deepxml~\citep{dahiya2021deepxml} and its variant
(i.e., \decaf~\citep{mittal2021decaf}, \galaxc~\citep{saini2021galaxc}, \eclare~\citep{mittal2021eclare})
pre-train MLP encoders on \xmc sub-problems induced by label clusters.
They freeze the pre-trained word embedding and learn another MLP layer
with hard negative labels from HNSW~\citep{malkov2020hnsw}.
Notably, shallow embedding-based methods only show competitive
performance on short-text \xmc problems where the number of input tokens is small.

To better handle longer text sequence, \attentionxml~\citep{you2019attentionxml}
uses BiLSTMs and label-aware attention as the scoring function.
For better scalability to large output spaces, training of \attentionxml involves various
negative sampling strategies to avoid back-propagating the entire label embedding layer.
More recently, \lightxml~\citep{jiang2021lightxml} adopts the transformer models as text 
encoder, and performs label shortlist and re-ranking with the same transformer encoder.
By capturing rich semantic information from input text, \lightxml 
establishes competitive results on public \xmc benchmarks.  

\noindent \paragraph{\bf{Comparing \pecos with \attentionxml.}}
\pecos induces two neural-based realizations using Transformer encoders,
which are \xtransformer~\citep{chang2020xmctransformer} and \xrtransformer~\citep{zhang2021fast}.
There are three main differences between \xrtransformer and \attentionxml.
First, \xrtransformer captures better semantic embeddings for long text sequence using Transformers.
Second, \xrtransformer can easily leverage any pre-trained Transformer models from the literature.
Finally, \xrtransformer is optimized with cost-sensitive loss induced by recursive course-to-fine signals.

\section{Experimental Results}
\label{sec:exp}
In this section, we compare various realization of \pecos with recent \xmc models on six real-world extreme multi-label text
classification datasets: \eurlex, \wikis, \amzcat, \wikil, \amzsmall and \amzlarge.
%These datasets represent some of the largest publicly available datasets in terms of number of labels, for example, \amzlarge has about 2.8M labels.
Details of these datasets and its statistics are presented in Table~\ref{tab:data}.
We use the same raw text input, sparse feature representations, and training/test data split as in
~\citet{you2019attentionxml,chang2020xmctransformer,zhang2021fast,jiang2021lightxml} to have a fair and reproducible comparison.

% data stats
\begin{table*}[!t]
    \centering
    \caption{Data Statistics.
      $n_{\text{trn}}, n_{\text{tst}}$ refer to the number of instances in the training and test sets, respectively.
      $\abs{\cD_{\text{trn}}}, \abs{\cD_{\text{tst}}}$ refer to the number of word tokens in the training and test corpus, respectively.
      $d$ is the dimension of \tfidf feature vector.
      $L$ is the number of labels, $\bar{L}$ the average number of labels per
      instance, $\bar{n}$ the average number of instances per label.
      These six publicly available benchmark datasets
      are downloaded from \url{https://github.com/yourh/AttentionXML} which
      are the same as \attentionxml~\citep{you2019attentionxml} for fair comparison.
    }
    \label{tab:data}
    \resizebox{1.0\textwidth}{!}{
    \begin{tabular}{c|rrrrrrrr}
    \toprule
    Dataset            & $n_{\text{trn}}$ & $n_{\text{tst}}$ & $\abs{\cD_{\text{trn}}}$ &  $\abs{\cD_{\text{tst}}}$ & $d$       & $L$       & $\bar{L}$ & $\bar{n}$ \\
    \midrule
        \eurlex         &       15,449    &           3,865  &              19,166,707  &                 4,741,799 &   186,104 &     3,956 &     5.30  &     20.79 \\
        \wikis          &       14,146    &           6,616  &              29,603,208  &                13,513,133 &   101,938 &    30,938 &    18.64  &      8.52 \\
        \amzcat         &    1,186,239    &         306,782  &             250,940,894  &                64,755,034 &   203,882 &    13,330 &     5.04  &    448.57 \\
        \wikil          &    1,779,881    &         769,421  &           1,463,197,965  &               632,463,513 & 2,381,304 &   501,070 &     4.75  &     16.86 \\
        \amzsmall       &      490,449    &         153,025  &             119,981,978  &                36,509,660 &   135,909 &   670,091 &     5.45  &      3.99 \\
        \amzlarge       &    1,717,899    &         742,507  &             174,559,559  &                75,506,184 &   337,067 & 2,812,281 &    36.04  &     22.02 \\
    \bottomrule
    \end{tabular}
    }
\end{table*}

% main results
\begin{table*}[!h]
    \centering
    \caption{Comparison of \xrlinear, \xtransformer and \xrtransformer
    with recent \xmc methods on six publicly available datasets.
    Results of non-\pecos models are taken from~\citet[Table 3]{you2019attentionxml} and~\citet[Table 2]{jiang2021lightxml}.
    The results show that \xrtransformer achieves state-of-the-art precision numbers.
    }
    %\resizebox{0.90\textwidth}{!}{
    \resizebox{1.0\textwidth}{!}{
    \begin{tabular}{cccccccc}
        \toprule
        Methods & Prec@1 & Prec@3 & Prec@5 & Methods & Prec@1 & Prec@3 & Prec@5 \\
        \hline
        \hline
        \multicolumn{4}{c}{ \eurlex }                             & \multicolumn{4}{c}{ \wikis } \\
        \midrule
        \annexml          & 79.66 & 64.94 & 53.52 & \annexml          & 86.46 & 74.28 & 64.20 \\
        \discmec          & 83.21 & 70.39 & 58.73 & \discmec          & 84.13 & 74.72 & 65.94 \\
        \pfastrexml       & 73.14 & 60.16 & 50.54 & \pfastrexml       & 83.57 & 68.61 & 59.10 \\
        \parabel          & 82.12 & 68.91 & 57.89 & \parabel          & 84.19 & 72.46 & 63.37 \\
        \xtext            & 79.17 & 66.80 & 56.09 & \xtext            & 83.66 & 73.28 & 64.51 \\
        \bonsai           & 82.30 & 69.55 & 58.35 & \bonsai           & 84.52 & 73.76 & 64.69 \\
        %\mlcseq           & 62.77 & 59.06 & 51.32 & \mlcseq           & 80.79 & 58.59 & 54.66 \\
        \fasttext         & 71.59 & 60.51 & 51.07 & \fasttext         & 82.26 & 65.93 & 55.25 \\
        \xmlcnn           & 75.32 & 60.14 & 49.21 & \xmlcnn           & 81.41 & 66.23 & 56.11 \\
        \attentionxml     & 87.12 & 73.99 & 61.92 & \attentionxml     & 87.47 & 78.48 & 69.37 \\
        \lightxml         & 87.63 & 75.89 & \best{63.36} & \lightxml & \best{89.45} & 78.96 & 69.85 \\
        \midrule
        \xrlinear         & 82.07 & 69.61 & 58.23 & \xrlinear         & 84.55 & 73.02 & 64.24 \\
        \xtransformer     & 87.61 & 75.39 & 63.05 & \xtransformer     & 88.26 & 78.51 & 69.68 \\
        \xrtransformer    & \best{88.41} & \best{75.97} & {63.18} &
        \xrtransformer    & {88.69} & \best{80.17} & \best{70.91} \\
        \hline
        \hline
        \multicolumn{4}{c}{ \amzcat }     & \multicolumn{4}{c}{ \wikil } \\
        \midrule
        \annexml          & 93.54 & 78.36 & 63.30 & \annexml          & 64.22 & 43.15 & 32.79 \\
        \discmec          & 93.81 & 79.08 & 64.06 & \discmec          & 70.21 & 50.57 & 39.68 \\
        \pfastrexml       & 91.75 & 77.97 & 63.68 & \pfastrexml       & 56.25 & 37.32 & 28.16 \\
        \parabel          & 93.02 & 79.14 & 64.51 & \parabel          & 68.70 & 49.57 & 38.64 \\
        \xtext            & 92.50 & 78.12 & 63.51 & \xtext            & 65.17 & 46.32 & 36.15 \\
        \napkinxc         & 93.04 & 78.44 & 63.70 & \napkinxc         & 66.77 & 47.63 & 36.94 \\
        \bonsai           & 92.98 & 79.13 & 64.46 & \bonsai           & 69.26 & 49.80 & 38.83 \\
        %\mlcseq           & 94.26 & 69.45 & 57.55 & \mlcseq           & - & - & - \\
        \fasttext         & 90.55 & 77.36 & 62.92 & \fasttext         & 31.59 & 18.47 & 13.47 \\
        \xmlcnn           & 93.26 & 77.06 & 61.40 & \xmlcnn           & - & - & - \\
        \attentionxml     & 95.92 & 82.41 & 67.31 & \attentionxml     & 76.95 & 58.42 & 46.14 \\
        \lightxml         & 96.77 & \best{84.02} & \best{68.70} & \lightxml & 77.78 & 58.85 & 45.57 \\
        \midrule
        \xrlinear         & 92.97 & 78.94 & 64.30 & \xrlinear         & 68.12 & 49.07 & 38.39 \\
        \xtransformer     & 96.48 & 83.41 & 68.19 &\xtransformer & 77.09 & 57.51 & 45.28 \\
        \xrtransformer    & \best{96.79} & {83.66} & {68.04} &
        \xrtransformer    & \best{79.40} & \best{59.02} & \best{ 46.25 } \\
        \hline
        \hline
        \multicolumn{4}{c}{ \amzsmall }     & \multicolumn{4}{c}{ \amzlarge } \\
        \midrule
        \annexml          & 42.09 & 36.61 & 32.75 & \annexml          & 49.30 & 45.55 & 43.11 \\
        \discmec          & 44.78 & 39.72 & 36.17 & \discmec          & 47.34 & 44.96 & 42.80 \\
        \pfastrexml       & 36.84 & 34.23 & 32.09 & \pfastrexml       & 43.83 & 41.81 & 40.09 \\
        \parabel          & 44.91 & 39.77 & 35.98 & \parabel          & 47.42 & 44.66 & 42.55 \\
        \xtext            & 42.54 & 37.93 & 34.63 & \xtext            & 42.20 & 39.28 & 37.24 \\
        \napkinxc         & 43.54 & 38.71 & 35.15 & \napkinxc         & 46.23 & 43.48 & 41.41 \\
        \bonsai           & 45.58 & 40.39 & 36.60 & \bonsai           & 48.45 & 45.65 & 43.49 \\
        %\mlcseq           &   -   &   -   &   -   & \mlcseq           &   -   &   -   &   -   \\
        \fasttext         & 24.35 & 21.26 & 19.14 & \fasttext         & 22.51 & 19.05 & 16.99 \\
        \xmlcnn           & 33.41 & 30.00 & 27.42 & \xmlcnn           &   -   &   -   &   -   \\
        \attentionxml     & 47.58 & 42.61 & 38.92 & \attentionxml     & 50.86 & 48.04 & 45.83 \\
        \lightxml         & 49.10 & 43.83 & 39.85 & \lightxml         &   -   &   -   &   - \\
        \midrule
        \xrlinear         & 45.36 & 40.35 & 36.71 & \xrlinear         & 47.96 & 45.09 & 42.96 \\
        \xtransformer     & 48.07 & 42.96 & 39.12 & \xtransformer     & 51.20 & 47.81 & 45.07 \\
        \xrtransformer    & \best{50.11} & \best{44.56} & \best{40.64} &
        \xrtransformer    & \best{54.20} & \best{50.81} & \best{48.26} \\
        \bottomrule
    \end{tabular}
    }
    \label{tab:main-results}
\end{table*}

In Section~\ref{sec:exp-performance}, we focus on the performance of various
models and demonstrate that \xrtransformer, a realization of \pecos framework
with a recursive Transformer matcher, achieves state-of-the-art prediction performance.
In Section~\ref{sec:exp-infra-cost}, we show that \xrlinear (Section~\ref{sec:xrlinear}),
a linear counterpart of \xrtransformer, achieves satisfactory prediction performance while
requiring substantially less training time.
In Section~\ref{sec:exp-realtime}, we demonstrate that the efficiency of the inference procedure for \xrlinear,
which allows it to serve real-time requests.
Finally, in Section~\ref{sec:exp-ablation}, we present the ablation study of \xrlinear to examine
the effectiveness of semantic label clustering and model ensembling.

\subsection{Performance Comparison}
\label{sec:exp-performance}
To compare the predictive performance of various models,
we use the widely used precision and recall metrics for
the \xmc task~\citep{prabhu2014fastxml,bhatia2015sparse,jain2016extreme,prabhu2018parabel,reddi2018stochastic}.
In particular, for an input $\bx$ and the corresponding ground truth
$\by\in\cY$, the Prec@$p$ ($p=1,3,5$) and Recall@$p$ ($p=1,3,5$) for the top-$b$
predictions $f_b(\bx)$ are defined as follows:
\begin{equation*}
\text{Prec}@b = \frac{1}{b}\sum_{\ell \in f_b(\bx)} y_{\ell},
\quad
\text{Recall}@b = \frac{1}{\text{nnz}(\by)}\sum_{\ell \in f_b(\bx)} y_{\ell}.
\end{equation*}
We consider three \pecos instantiations:
\begin{itemize}
  \item \xrlinear:
    We use \pifa as label embeddings to construct the hierarchical label tree (HLT)
    with branching factor $B=32$ in Algorithm~\ref{alg:bisection-kmeans}.
    We use TFN as the negative sampling in Algorithm~\ref{alg:xrlinear}.
    Similar to \parabel~\citep{prabhu2018parabel},
    we use beam size $b=10$ and an ensemble of three HLTs for the prediction stage.
    %The number of label clusters $K$ for each dataset is given in Table~\ref{tab:data}.
  \item \xtransformer: \pecos with \textit{non-recursive} Transformer matchers.
    Predictions are an ensemble of 9 \xtransformer models with three encoders and three HLTs.
    Detailed specifications and hyperparameters can be found in \citet[Section 4.2]{chang2020xmctransformer}.
  \item \xrtransformer: \pecos with \textit{recursive} Transformer matchers.
    Predictions are an ensemble of 3 \xrtransformer models with three encoders.
    Detailed specifications and hyperparameters can be found in \citet[Section 5]{zhang2021fast}.
\end{itemize}
We then include the following \xmc models in our comparisons.
\begin{itemize}
  \item Embedding-based Approaches: \annexml~\citep{tagami2017annexml}
  \item OVR-based Approaches: \discmec~\citep{babbar2017dismec}
  \item Tree-based Approaches: \pfastrexml~\citep{jain2016extreme},
    \parabel~\citep{prabhu2018parabel}, \xtext~\citep{wydmuch2018no},
    \bonsai~\citep{khandagale2020bonsai}, and
    \napkinxc~\citep{jasinska2020probabilistic}.
  \item Deep learning-based Approaches: \fasttext~\citep{joulin2017bag}, 
    \xmlcnn~\citep{liu2017deep}, \attentionxml~\citep{you2019attentionxml} and \lightxml~\citep{jiang2021lightxml}.
\end{itemize}
Note that in terms of our three phase framework for PECOS, \parabel~\citep{prabhu2018parabel} may be interpreted as a special
case of~\xrlinear with the following special choices: \pifa label representation +
Algorithm~\ref{alg:bisection-kmeans} (with $B=2$) + TFN negative sampling scheme.
In other words, \xrlinear with $B=32$ in Table~\ref{tab:main-results} has a smaller depth of hierarchical label tree,
which enjoys faster inference time in practice.

Table~\ref{tab:main-results} shows that our proposed \xrtransformer method outperforms
other competitive deep learning based \xmc methods (e.g., \attentionxml and \lightxml) on most metrics, 
especially on datasets with large output spaces such as \amzsmall and \amzlarge.
This verifies the effectiveness of recursive learning of transformer encoders on large output space problems.
While \xtransformer and \xrtransformer results in better predictive performance compared to its linear counterpart \xrlinear,
they also requires considerably longer training time, as we will see in the Section~\ref{sec:exp-infra-cost}.

\begin{table}[t!]
  \caption{Training time (in seconds) versus predictive performance of various \pecos realizations.}
  \label{tab:comp-cost}
	\begin{resize}{0.90\linewidth}
    \centering
		\begin{tabular}{l|rrr|rrr|rr}
			\toprule
      \toprule
			\multicolumn{7}{c}{\eurlex ($\abs{\cY}=3,956,\ n_{\text{trn}}=15,449,\ n_{\text{tst}}=3,865$)} & \multicolumn{2}{c}{Model Training}\\
			                                 &     Prec@1 &      Prec@3&     Prec@5 &    Recall@1&    Recall@3&    Recall@5&  Time (s)      \\
	  \midrule
      \xrtransformer                   &\best{88.41}&\best{75.97}&\best{63.18}&\best{17.93}&\best{45.26}&\best{61.49}&    2,880.0     \\
      \xtransformer						         &     {87.61}&     {75.39}&     {63.05}&     {17.78}&     {44.92}&     {61.35}&   26,766.0     \\
			\xrlinear &\multicolumn{3}{c|}{}&\multicolumn{3}{c|}{}&\multicolumn{2}{c}{}\\
			$\quad$ TFN                      &     {82.07}&     {69.61}&     {58.23}&     {16.59}&     {41.36}&     {56.60}&\best{7.4}     \\
			$\quad$ TFN+MAN                  &     {83.08}&     {69.87}&     {58.18}&     {16.81}&     {41.55}&     {56.52}&     {20.2}    \\
			\midrule
			\midrule
			\multicolumn{7}{c}{\wikis ($\abs{\cY}=30,938,\ n_{\text{trn}}=14,146,\ n_{\text{tst}}=6,616$)} & \multicolumn{2}{c}{Model Training} \\
			                                 &     Prec@1 &      Prec@3&     Prec@5 &    Recall@1&    Recall@3&    Recall@5& Time (s)       \\
    \midrule % wiki10-31k
      \xrtransformer                   &\best{88.69}&\best{80.17}&\best{70.91}& \best{5.30}&\best{14.17}&\best{20.44}&     5.400.0    \\
      \xtransformer                    &     {88.26}&     {78.51}&     {69.68}&     { 5.28}&     {13.76}&     {19.79}&    51,815.0    \\
			\xrlinear &\multicolumn{3}{c|}{}&\multicolumn{3}{c|}{}&\multicolumn{2}{c}{}\\
			$\quad$ TFN                      &     {84.55}&     {73.02}&     {64.24}&     { 4.99}&     {12.72}&     {18.40}&\best{36.0}    \\
			$\quad$ TFN+MAN                  &     {84.70}&     {73.86}&     {64.76}&     { 5.02}&     {12.92}&     {18.57}&     {70.9}    \\
			\midrule
			\midrule
			\multicolumn{7}{c}{\amzcat ($\abs{\cY}=13,330,\ n_{\text{trn}}=1,186,239,\ n_{\text{tst}}=306,782$)} & \multicolumn{2}{c}{Model Training} \\
			                                 &     Prec@1 &      Prec@3&     Prec@5 &    Recall@1&    Recall@3&    Recall@5& Time (s)       \\
    \midrule % amazoncat-13k
      \xrtransformer                   &\best{96.79}&\best{83.66}&\best{68.19}&\best{27.69}&\best{63.31}&\best{79.37}&   47,520.0     \\
      \xtransformer                    &     {96.48}&     {83.41}&\best{68.19}&     {27.52}&     {63.11}&     {79.30}&    531,308.0   \\
			\xrlinear &\multicolumn{3}{c|}{}&\multicolumn{3}{c|}{}&\multicolumn{2}{c}{}\\
			$\quad$ TFN                      &      {93.06}&     {78.95}&     {64.28}&     {26.33}&     {59.78}&     {75.20}&\best{220.2}  \\
      $\quad$ TFN+MAN                  &      {93.06}&     {78.95}&     {64.20}&     {26.30}&     {59.77}&     {75.17}&    {1,074.3} \\
			\midrule
			\midrule
			\multicolumn{7}{c}{\wikil ($\abs{\cY}=501,070,\ n_{\text{trn}}=1,779,881,\ n_{\text{tst}}=769,421$)} & \multicolumn{2}{c}{Model Training} \\
			                                 &     Prec@1 &      Prec@3&     Prec@5 &    Recall@1&    Recall@3&    Recall@5& Time (s)       \\
    \midrule % wiki-500k
      \xrtransformer                   &\best{79.40}&\best{59.02}&\best{46.25}&\best{26.59}&\best{49.61}&\best{59.53}&   136,800.0    \\
    \xtransformer                      &     {77.09}&     {57.51}&     {45.28}&     {25.51}&     {48.03}&     {58.05}&  2,005,550.0   \\
      \xrlinear &\multicolumn{3}{c|}{}&\multicolumn{3}{c|}{}&\multicolumn{2}{c}{}\\
      $\quad$ TFN                      &      {68.12}&     {49.07}&     {38.39}&     {22.18}&     {40.72}&     {49.21}&\best{2,796.6} \\
      $\quad$ TFN+MAN                  &      {68.77}&     {48.24}&     {37.07}&     {22.57}&     {40.33}&     {47.87}&    {19,356.2} \\
			\midrule
			\midrule
			\multicolumn{7}{c}{\amzsmall ($\abs{\cY}=670,091,\ n_{\text{trn}}=490,449,\ n_{\text{tst}}=153,025$)} & \multicolumn{2}{c}{Model Training} \\
			                                 &     Prec@1 &      Prec@3&     Prec@5 &    Recall@1&    Recall@3&    Recall@5& Time (s)       \\
    \midrule % amazon-670k
      \xrtransformer                   &\best{50.11}&\best{44.56}&\best{40.64}&\best{10.52}&\best{26.02}&\best{38.29}&   37,800.0     \\
      \xtransformer                    &     {48.07}&     {42.96}&     {39.12}&     {9.94}&     {24.90}&     {36.71}& 1,853,263.0     \\
      \xrlinear &\multicolumn{3}{c|}{}&\multicolumn{3}{c|}{}&\multicolumn{2}{c}{}\\
      $\quad$ TFN                      &      {45.36}&     {40.35}&     {36.71}&     { 9.44}&     {23.43}&     {34.45}&\best{147.2}   \\
      $\quad$ TFN+MAN                  &      {45.81}&     {40.64}&     {36.82}&     { 9.63}&     {23.67}&     {34.61}&     {928.4}   \\
			\midrule
			\midrule
			\multicolumn{7}{c}{\amzlarge ($\abs{\cY}=2,812,281,\ n_{\text{trn}}=1,717,899,\ n_{\text{tst}}=742,507$)} & \multicolumn{2}{c}{Model Training} \\
			                                 &     Prec@1 &      Prec@3&     Prec@5 &    Recall@1&    Recall@3&    Recall@5& Time (s)       \\
    \midrule % amazon-3m
      \xrtransformer                   &\best{54.20}&\best{50.81}&\best{48.26}&\best{3.93}&\best{9.59}&\best{13.99}&   105,480.0      \\
     \xtransformer                     &     {51.20} &     {47.81}&     {45.07}&     {3.28}&     {8.03}&     {11.65}&  {1,951,324.0}  \\
      \xrlinear &\multicolumn{3}{c|}{}&\multicolumn{3}{c|}{}&\multicolumn{2}{c}{}\\
      $\quad$ TFN                      &      {47.96}&     {45.09}&     {42.96}&     { 3.04}&     { 7.54}&     {11.12}&\best{1,453.5} \\
      $\quad$ TFN+MAN                  &      {48.64}&     {45.90}&     {43.76}&     { 3.28}&     { 8.05}&     {11.81}&    { 4,971.6} \\
			\bottomrule
      \bottomrule
		\end{tabular}
	\end{resize}
\end{table}

\subsection{Training Time versus Predictive Performance}
\label{sec:exp-infra-cost}
In this section, we analyze various \pecos realizations based on their training time and predictive performance.
%We compare various \pecos realizations with another popular text classification package \fasttext~\citep{joulin2017bag}.
All the experiments of \xrlinear are run on a r5.24xlarge AWS instance,
which contains 96 Intel Xeon Platinum 8000 CPUs and 768 GB RAM.
All the experiments of \xtransformer and \xrtransformer, are obtained on a p3.16xlarge AWS instance,
which contains 8 Nvidia V100 GPUs.

\begin{comment}
\paragraph{\bf Compared Methods.}
\begin{itemize}
  \item \xrlinear:
    \begin{itemize}
      \item \url{https://code.amazon.com/packages/MidasPecos/trees/mainline}.
      \item Text Vectorizer: we use a \tfidf text vectorizer with bigrams
      \item Semantic Label Repression: \pifa.
      \item Semantic Label Indexing: spherical K-means with bisections with default parameters.
      \item Machine Learning Matcher and Ranker: \xrlinear implementation described in
        Algorithm~\ref{alg:xrlinear} with default parameters.
    \end{itemize}
  \item \fasttext:
    \begin{itemize}
      \item \url{https://github.com/facebookresearch/fastText/releases/tag/v0.9.1}.
      \item We use the option to enable bigrams in \fasttext
      \item Two loss functions are tested: softmax, and hierarchical softmax.
        Note that the negative sampling and one-versus-reset options give much
        worse results than softmax and hierarchical softmax.
      \item The default values of the learning rate and epoch in\fasttext
        perform extremely poorly. We perform simple grid search to use $0.5$ as
        the learning rate and 200 as the number of epochs.
    \end{itemize}
\end{itemize}
\paragraph{Results.}
\end{comment}

Experimental results are shown in Table~\ref{tab:comp-cost}, which include
two variants of \xrlinear with different negative mining (i.e., TFN, TFN+MAN).
We can clearly see that \xrlinear is the most efficient approach in terms of training time,
followed by \xrtransformer and then \xtransformer.
In particular, \xrlinear with TFN negative sampling is often 2x to 5x faster than
\xrlinear with TFN+MAN sampling, because the latter requires model inference on the
large training set to generate hard negative based on the model parameters.
Nevertheless, \xrlinear with TFN+MAN may lead to better predictive performance on
larger output space datasets such as \amzlarge.

Compared to \xrlinear, on the other hand, \xtransformer and \xrtransformer
yield state-of-the-art precision and recall results at the cost of larger training time,
where \xrtransformer is often 10x faster than \xtransformer.
This verifies the effectiveness of recursive training for Transformer matchers on the large output space datasets.  
It is noteworthy that \pecos is
flexible to have realizations like \xrtransformer which yields the
best prediction performance and realizations like \xrlinear which
strikes a good balance between prediction performance and training cost.
Note that even though relatively cheaper compared with \xtransformer, the 
\xrtransformer method still requires faster and more expensive GPU hardware.
This flexibility allows practitioners to choose the most appropriate \pecos model for their applications.

\subsection{Real-Time Inference}
\label{sec:exp-realtime}
Next, we compare the real-time inference latency of various \pecos realization with competitive \xmc methods.
Specifically, in real-time mode, we consider the test instances are fed one-by-one to the model.
Table~\ref{tab:exp-realtime-parabel} compares the inference latency (milliseconds per input instance)
among \parabel, \xrlinear, \napkinxc, \xtransformer and \xrtransformer in real time mode.
Real-time experiments for \parabel, \napkinxc and \xrlinear are conducted on a AWS instance r5.4xlarge
using \textit{single} thread while \xtransformer and \xrtransformer are evaluated on a Tesla V100 GPU.
In our experiments, we randomly sampled 10,000 test inputs/instances for reporting the numbers in Table~\ref{tab:exp-realtime-parabel}.

We implemented our version of \parabel and \napkinxc for fair comparison in the real-time inference case.
The original \parabel code is designed for batch input mode, so it is slow in real-time mode.
\parabel, \xrlinear, \napkinxc all use the same model parameters and same number of branch splits, $B=32$.
\xrlinear and \napkinxc use hash method to look up the non-empty rows.
Beam size and topk are both set to 10 for all experiments in Table~\ref{tab:exp-realtime-parabel}.

As we can see, \xrlinear is much faster than \napkinxc and \parabel across all datasets.
\xtransformer and \xrtransformer are deep learning models which achieve better precision and recall
but require much larger inference time due to expensive transformer encoders. 

\begin{table}[t!]
	\begin{resize}{1\linewidth}

    \centering
    \caption{Online Inference Latency (milliseconds per input) for \parabel, \napkinxc, \xrlinear, 
    \xrtransformer and \xtransformer. \xtransformer and \xrtransformer are evaluated on a Nvidia Tesla
    V100 GPU while other models are evaluated on an AWS instance r5.4xlarge using a single thread. }
    \label{tab:exp-realtime-parabel}
	\begin{tabular}{r | r | r | r | r | r |r }
    	\toprule
             			&\eurlex & \wikis & \amzcat & \wikil & \amzsmall & \amzlarge \\
		\midrule
		\parabel        &   1.20 &   5.77 &  17.00 & 175.00 &  10.70 &  44.60 \\
    \napkinxc       &   1.63 &   7.10 &   1.93 &  12.60 &   2.80 &   3.59 \\
		\xrlinear       & \best{0.20} & \best{1.06} & \best{0.33} & \best{2.26} & \best{0.48} & \best{0.61} \\
		\xtransformer 	& 433.87 & 433.20 & 428.58 & 433.33 & 432.24 & 451.97 \\
		\xrtransformer 	&  66.90 & 117.30 &  78.30 & 101.70 &  92.70 & 105.60 \\
  	\bottomrule
	\end{tabular}
	\end{resize}
\end{table}

\subsection{Ablation Study of \xrlinear}
\label{sec:exp-ablation}
In Table~\ref{tab:ablation-xrl}, we compare different configurations of hierarchical label tree (HLT) as the ablation study of \xrlinear.
The experiment results are conducted on an r5.24xlarge AWS instance, which contains 96 Intel Xeon Platinum 8000 CPUs and 768 GB RAM.
Note that we use the multi-threading batch-mode for model predictions and report the total prediction time in seconds.

First, we investigate different tree depths of HLT by vary branching splits $B=\{2, 8, 32\}$. 
From Table~\ref{tab:ablation-xrl}, we observed that $B=32$ (shallower HLTs) usually results in better predictive performance
compared to $B=2$ (deeper HLTs).
Furthermore, the prediction time of shallower HLTs are also faster than the prediction time of deeper HLTs,
which justifies the default choice of $B=32$ for experiments of all previous sections.
We also explore a randomly clustered HLT on \eurlex, where
the Prec@k drops to 79.3, 67.1, and 55.9, for $k=1,3,5$, respectively.
Given this significant drop, we omit the results of random clusters in Table~\ref{tab:ablation-xrl}.

Finally, we compare the performance versus training time of a single HLT ($T=1$) versus an ensemble of three HLTs ($T=3$).
From Table~\ref{tab:ablation-xrl}, \xrlinear with $T=3$ has better precision and recall compared to \xrlinear with $T=1$.
However, it also comes with the price of larger training and prediction time.

\begin{table}[t!]
  \caption{
    Ablation study of \xrlinear with respective to different configurations of the hierarchical label tree (HLT).
    $T$ is the number of HLT, where $T=1$ refers to \xrlinear with a single HLT
    while $T=3$ refers to \xrlinear with an ensemble of three HLTs.
    $B$ and $D$ are the branching splits and depth of HLT, respectively.
    Training time and prediction time (in seconds) are measured on an r5.24xlarge AWS instance,
    which contains 96 Intel Xeon Platinum 8000 CPUs and 768 GB RAM.
    Note that we use TFN sampling to train the model and consider beam size $b=10$ for the model prediction
    using multi-threading batch-mode.
  }
  \label{tab:ablation-xrl}
  \begin{resize}{1.0\linewidth}
    \begin{tabular}{rrr|rrr|rrr|rr}
      \toprule
      \toprule
			\multicolumn{11}{c}{\eurlex ($\abs{\cY}=3,956,\ n_{\text{trn}}=15,449,\ n_{\text{tst}}=3,865$)} \\
                  $T$ & $B$ & $D$ & Prec@1  & Prec@3  & Prec@5  & Recall@1  & Recall@3  & Recall@5  & Train Time  & Predict Time \\
      \midrule
                   3  &   2  &  7 & 81.47 & 69.13 & 57.87 & 16.51 & 41.10 & 56.28 &     8.70 &   0.28 \\
                   3  &   8  &  3 & 81.79 & 69.30 & 58.04 & 16.54 & 41.16 & 56.46 &     7.40 &   0.32 \\
                   3  &  32  &  3 & 82.07 & 69.61 & 58.23 & 16.59 & 41.36 & 56.60 &     7.47 &   0.28 \\
                   1  &  32  &  3 & 82.48 & 68.83 & 57.61 & 16.65 & 40.85 & 55.98 &     2.49 &   0.09 \\
      \midrule
      \midrule
      \multicolumn{11}{c}{\wikis ($\abs{\cY}=30,938,\ n_{\text{trn}}=14,146,\ n_{\text{tst}}=6,616$)} \\
                  $T$ & $B$ & $D$ & Prec@1  & Prec@3  & Prec@5  & Recall@1  & Recall@3  & Recall@5  & Train Time  & Predict Time \\
      \midrule
                   3  &  2  & 10 & 84.19 & 72.57 & 63.39 &  4.97 & 12.61 & 18.12 &    36.05 &   1.20 \\
                   3  &  8  &  4 & 84.40 & 72.87 & 64.09 &  4.99 & 12.68 & 18.33 &    30.38 &   0.65 \\
                   3  & 32  &  3 & 84.55 & 73.02 & 64.24 &  4.99 & 12.72 & 18.40 &    31.18 &   0.65 \\
                   1  & 32  &  3 & 84.14 & 72.85 & 64.09 &  4.97 & 12.69 & 18.35 &    10.39 &   0.22 \\
      \midrule
      \midrule
      \multicolumn{11}{c}{\amzcat ($\abs{\cY}=13,330,\ n_{\text{trn}}=1,186,239,\ n_{\text{tst}}=306,782$)} \\
                  $T$  & $B$ & $D$ & Prec@1  & Prec@3  & Prec@5  & Recall@1  & Recall@3  & Recall@5  & Train Time  & Predict Time \\
      \midrule
                   3   &  2  &  9 & 93.06 & 78.95 & 64.28 & 26.33 & 59.78 & 75.20 &   220.20 &  22.52 \\
                   3   &  8  &  4 & 93.01 & 78.93 & 64.28 & 26.31 & 59.74 & 75.17 &   144.61 &  14.96 \\
                   3   & 32  &  3 & 92.97 & 78.94 & 64.30 & 26.28 & 59.74 & 75.20 &   134.00 &  15.11 \\
                   1   & 32  &  3 & 92.53 & 78.45 & 63.85 & 26.13 & 59.38 & 74.72 &    44.67 &   5.04 \\
      \midrule
      \midrule
      \multicolumn{11}{c}{\wikil ($\abs{\cY}=501,070,\ n_{\text{trn}}=1,779,881,\ n_{\text{tst}}=769,421$)} \\
                  $T$  & $B$ & $D$ & Prec@1  & Prec@3  & Prec@5  & Recall@1  & Recall@3  & Recall@5  & Train Time  & Predict Time \\
      \midrule
                   3   &  2  & 14 & 68.44 & 49.28 & 38.58 & 22.30 & 40.89 & 49.44 & 2,933.10 & 177.16 \\
                   3   &  8  &  6 & 68.27 & 49.19 & 38.50 & 22.23 & 40.81 & 49.33 & 2,678.91 &  98.73 \\
                   3   & 32  &  4 & 68.12 & 49.07 & 38.39 & 22.18 & 40.72 & 49.21 & 2,796.64 &  92.38 \\
                   1   & 32  &  4 & 66.60 & 47.67 & 37.19 & 21.62 & 39.43 & 47.50 &   932.21 &  30.79 \\
      \midrule
      \midrule
      \multicolumn{11}{c}{\amzsmall ($\abs{\cY}=670,091,\ n_{\text{trn}}=490,449,\ n_{\text{tst}}=153,025$)} \\
                  $T$ & $B$ & $D$ & Prec@1  & Prec@3  & Prec@5  & Recall@1  & Recall@3  & Recall@5  & Train Time  & Predict Time \\
      \midrule
                   3  &  2  & 14 & 45.05 & 40.09 & 36.47 &  9.37 & 23.25 & 34.20 &   155.61 &  16.56 \\
                   3  &  8  &  6 & 45.35 & 40.26 & 36.59 &  9.43 & 23.36 & 34.31 &   145.25 &  11.77 \\
                   3  & 32  &  4 & 45.36 & 40.35 & 36.71 &  9.44 & 23.43 & 34.45 &   147.23 &  11.52 \\
                   1  & 32  &  4 & 44.14 & 39.06 & 35.30 &  9.17 & 22.63 & 33.07 &    49.08 &   3.84 \\
      \midrule
      \midrule
      \multicolumn{11}{c}{\amzlarge ($\abs{\cY}=2,812,281,\ n_{\text{trn}}=1,717,899,\ n_{\text{tst}}=742,507$)} \\
                  $T$ & $B$ & $D$ & Prec@1  & Prec@3  & Prec@5  & Recall@1  & Recall@3  & Recall@5  & Train Time  & Predict Time \\
      \midrule
                   3  &  2  & 16 & 47.30 & 44.38 & 42.23 &  2.96 &  7.32 & 10.79 & 1,481.56 &  99.74 \\
                   3  &  8  &  6 & 47.65 & 44.81 & 42.68 &  3.00 &  7.45 & 10.98 & 1,397.83 &  64.63 \\
                   3  & 32  &  4 & 47.96 & 45.09 & 42.96 &  3.04 &  7.54 & 11.12 & 1,453.53 &  63.82 \\
                   1  & 32  &  4 & 46.76 & 43.87 & 41.76 &  2.91 &  7.23 & 10.68 &   484.51 &  21.27 \\
      \bottomrule
    \end{tabular}    
  \end{resize}
\end{table}

\section{Conclusions and Future Work}
\label{sec:future}

In this paper, we have proposed \pecos, a versatile and modular machine learning framework
for solving prediction problems for very large output spaces. The flexibility
of \pecos allows practitioners to evaluate the trade-offs between performance
and training cost to identify the most appropriate \pecos variant for
their applications. In particular, we propose \xrlinear, a recursive
realization of our three-stage framework, which is highly efficient in both
training as well as inference, while being much less expensive in
training costs but yielding slightly lower quality than \xrtransformer,
which is a recursive neural realization of \pecos that yields state-of-the-art
prediction performance.

As future work, we plan to extend \pecos in various directions. One direction is to
explore more alternatives for each stage of \pecos such that it offers more
options to practitioners so they can identify the most appropriate variants for
their applications. For example, there are other
semantic label indexing strategies which might be promising alternatives, for example,
overlapping label clustering or approximate nearest neighbor
search schemes. In addition to linear rankers, we plan to explore more
sophisticated ranker choices such as gradient boosting models or neural network
models. To further improve the scalability of \pecos, we plan to use
distributed computation.
Another direction is to extend \pecos to handle
infinite output spaces that have structure. In particular,
we plan to conduct research to develop \pecos models that are able to not only
identify relevant labels from a given finite and large label set but also generate
relevant new labels when there is a generative model for these labels.
To facilitate this work by the research community,
we have open-sourced the \pecos software, which is available at 
\url{https://libpecos.org}.

\section*{Acknowledgement}
We thank Amazon for supporting this work. We also thank Lexing Ying, Philip Etter, and Tavor Baharav
for providing feedback on the manuscript.

%\newpage
\bibliographystyle{plainnat}
\bibliography{midas}

%\clearpage
% \newpage
% \appendix
% \input{amazon.tex}
\end{document}